\documentclass[10pt,twocolumn,letterpaper]{article}

\usepackage{iccv}
\usepackage{times}
\usepackage{epsfig}
\usepackage{graphicx}
\usepackage{amsmath}
\usepackage{amssymb}
\usepackage{booktabs}

\usepackage{subcaption}

\usepackage{color, colortbl}

\usepackage[misc]{ifsym}
\newcommand\blfootnote[1]{
  \begingroup
  \renewcommand\thefootnote{}\footnote{#1}
  \addtocounter{footnote}{-1}
  \endgroup
}

\usepackage{algorithm}
\usepackage{listings}
\usepackage{caption}
\usepackage{float}
\definecolor{DGray}{gray}{0.45}
\definecolor{Gray}{gray}{0.9}
\definecolor{codeblue}{rgb}{0.25,0.5,0.5}
\definecolor{codekw}{rgb}{0.85, 0.18, 0.50}
\definecolor{codekwb}{rgb}{0.0, 0.0, 1.00}
\floatstyle{plain}
\newfloat{Code}{htbp}{Code}
\restylefloat*{Code}
\floatname{Code}{Code}

\usepackage[pagebackref=true,breaklinks=true,letterpaper=true,colorlinks,bookmarks=false]{hyperref}
\usepackage[capitalize]{cleveref}
\crefname{section}{Sec.}{Secs.}
\Crefname{section}{Section}{Sections}
\Crefname{table}{Table}{Tables}
\crefname{table}{Tab.}{Tabs.}

\iccvfinalcopy 

\def\iccvPaperID{5666} 

\ificcvfinal\fi

\begin{document}

\title{StageInteractor: Query-based Object Detector with Cross-stage Interaction}

\author{
  Yao Teng\textsuperscript{1} \quad Haisong Liu\textsuperscript{1} \quad Sheng Guo\textsuperscript{3} \quad Limin Wang\textsuperscript{1,2,~\Letter} \\
  \textsuperscript{1}State Key Laboratory for Novel Software Technology, Nanjing University, China \\
  \textsuperscript{2}Shanghai AI Lab, China  \quad \textsuperscript{3}MYbank, Ant Group, China \\
}

\maketitle
\ificcvfinal\thispagestyle{empty}\fi

\begin{abstract}
Previous object detectors make predictions based on dense grid points or numerous preset anchors. Most of these detectors are trained with one-to-many label assignment strategies. On the contrary, recent query-based object detectors are based a sparse set of learnable queries refined by a series of decoder layers. The one-to-one label assignment is independently applied on each layer for deep supervision during training. Despite the great success of query-based object detection, however, this vanilla one-to-one label assignment strategy requires the detectors to have strong fine-grained discrimination and modeling capacity. In this paper, we propose a new query-based object detector with cross-stage interaction, coined as StageInteractor. During the forward pass, we come up with an efficient way to improve this modeling ability by reusing dynamic operators with lightweight adapters. As for the label assignment, a cross-stage label assigner is designed to improve the one-to-one label assignment. With this assigner, the training target class labels are gathered across stages and then reallocated to proper predictions at each decoder layer. On MS COCO benchmark, our model improves the baseline counterpart by 2.2 AP, and achieves a 44.8 AP with ResNet-50 as backbone, 100 queries and 12 training epochs. With longer training time and 300 queries, StageInteractor achieves 51.3 AP and 52.7 AP with ResNeXt-101-DCN and Swin-S, respectively. The code and models are made available at \url{https://github.com/MCG-NJU/StageInteractor}.
\end{abstract}
\blfootnote{\Letter: Corresponding author (lmwang@nju.edu.cn).}

\section{Introduction}
Object detection is a fundamental task in computer vision and acts as a cornerstone for many downstream tasks~\cite{sgg1,cong_stdysgg}.
It aims to localize and categorize all object instances in an image.
Over the past few decades, dense spatial prior has been widely applied in various detectors. These detectors make predictions based on either a large quantity of pre-defined anchors covering the whole image~\cite{fastrcnn,fasterrcnn,cascadercnn,ssd,focalloss,yolov2} or dense grid points in the feature map of this image~\cite{cornernet,cornercenternet,centernet,fcos,yolov1,reppoints}.
To deliver supervision signals to these detectors, most works employ the \textit{one-to-many label assignment} strategy~\cite{atss,autoassign,paa_assign,ot_assign,mutual_assign,anchorfree_assign_eccv22}~(\ie, the classification label and localization target of one ground-truth object could be assigned to multiple object predictions).
Although this paradigm is widely used in object detection, it suffers from redundant and near-duplicate predictions due to such label assignment~\cite{what_spz}, and thus relies heavily on specific post-processing algorithms for duplicate removal~\cite{softnms,relationnet,nms2}.

\begin{figure}[t]
  \centering
  \includegraphics[width=0.99\linewidth]{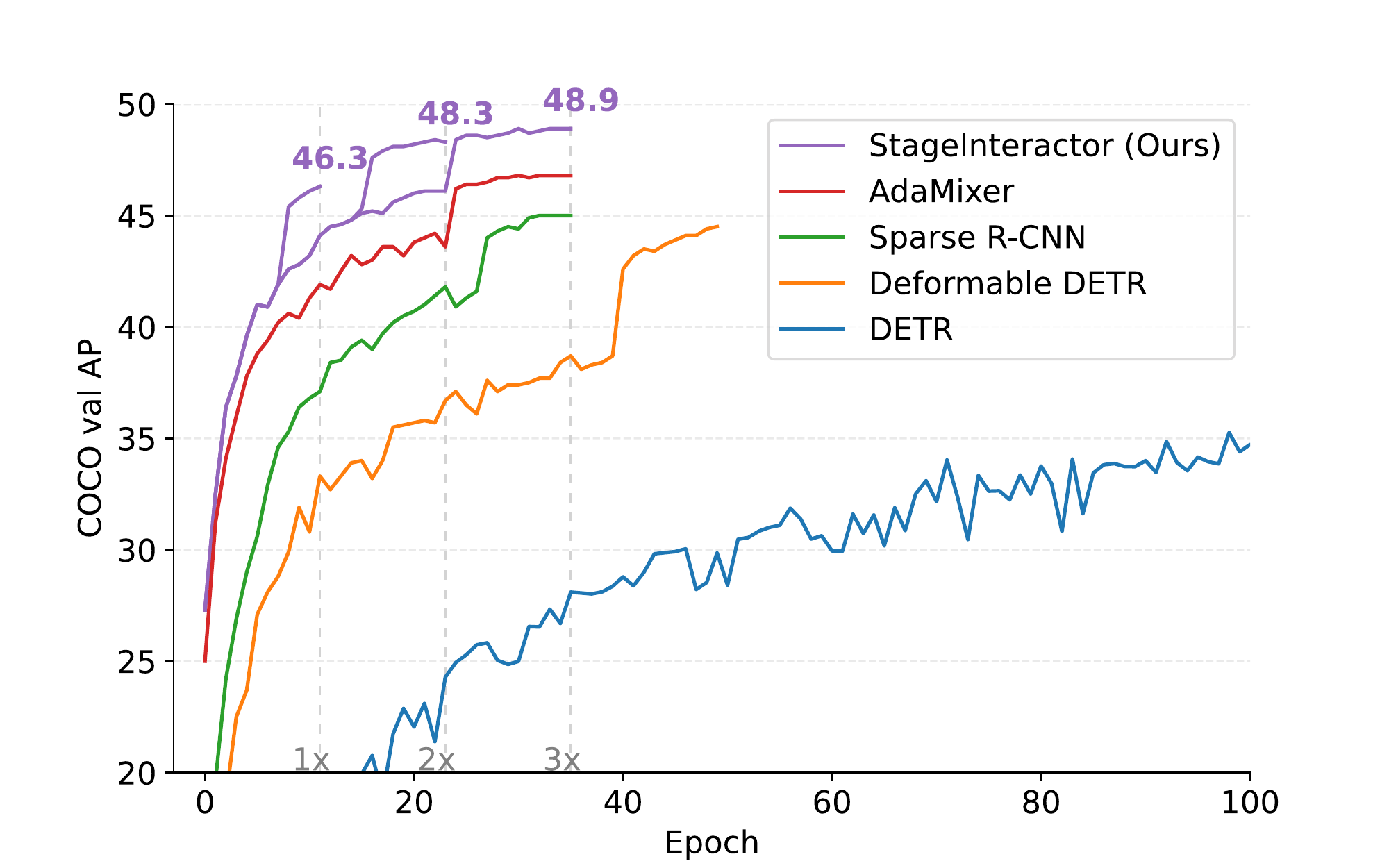}
   \caption{Convergence curves of our model and other query-based object detectors~\cite{detr, deformabledetr, sparsercnn, adamixer} with ResNet-50~\cite{resnet} on MS COCO~\cite{coco} minival set.}
   \label{fig:convergence}
\end{figure}

Recently, DETR~\cite{detr} and its variants~\cite{deformabledetr,sparsercnn,adamixer,conditionaldetrv2,dabdetr,smcadetr,boxer,glimpsedetr} open a new area of object detection.
These query-based object detectors get rid of the dense prior and view object detection as a set prediction problem. Specifically, they use a sparse set of \textit{queries} (\ie, learnable embeddings) to progressively capture the characteristics and location of objects with the help of a series of \textit{decoder layers}.
In each layer, image features are sampled and fused into the input queries via attention-like operation~\cite{transformer,deformabledetr} or dynamic mixing~\cite{sparsercnn,adamixer}. 
Then, the transformed queries are decoded into object predictions and also serve as the inputs of next layer.
As for the training of this paradigm, a kind of \textit{one-to-one label assignment} (\ie, each ground-truth object has to be assigned with only one prediction), termed as \textit{bipartite matching}, is independently adopted on each decoder layer for deep supervision~\cite{contrastive_deep_super,auxlossdetr,deepsuper}.
For inference, only the high-confidence outputs from the last layer are taken for evaluation.

\begin{figure}[t]
  \centering
  \includegraphics[width=0.99\linewidth]{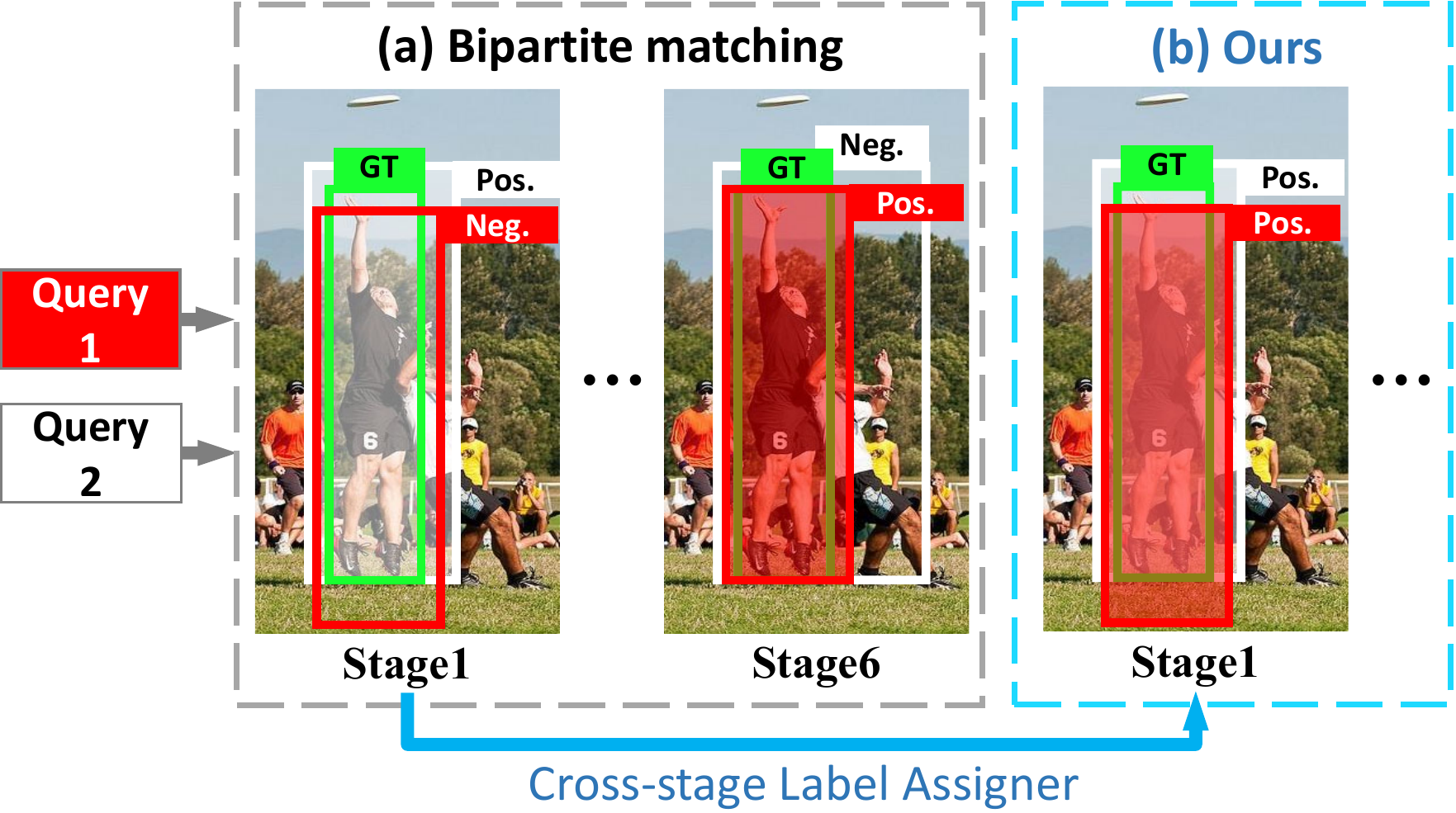}
   \caption{The results of label assignment at various stages. The green box denotes the ground-truth object \texttt{Person}. The red and white boxes denote object prediction derived from two different queries. \texttt{Pos} and \texttt{Neg} denote the positive sample and the negative sample, respectively. (a)~The white box is assigned with the ground-truth object \texttt{Person} by bipartite matching at the first stage, while the red box is not. But the opposite is true for the sixth stage. (b)~With our cross-stage label assigner, the red box in the first stage can be assigned with the ground-truth \texttt{Person}.
   }
  \vspace{-1.0em}
  \label{fig:la_example}
\end{figure}

However, this one-to-one label assignment requires the detectors to have strong fine-grained discrimination and modeling capacity.
On one hand, this strict bipartite matching would ask the detector to capture details to distinguish the predictions.
For example, as shown in~\cref{fig:la_example}\textcolor{red}{a}, 
although the predicted boxes of a query (colored in red) cover most of the ground-truth object (\texttt{Person}) at each decoder layer,
this object is assigned to the boxes of \textit{different} queries at \textit{different} stages (\texttt{Stage1}, white box; \texttt{Stage6}, red box).
In other words, only one of the predicted boxes (red or white) can become the positive sample at each stage.
To distinguish these boxes, the detector needs strong fine-grained discriminability to extract high-quality features from the image.
Unfortunately, this goal is hard to be fully accomplished.
As a result, we are motivated to directly modify the supervision of each decoder layer, \ie, introducing additional supervision from \textit{other stages} to assist the training of this layer, shown in~\cref{fig:la_example}\textcolor{red}{b}. 
In addition, the large modeling capacity is vital to the fine-grained discrimination.
To \textit{efficiently} improve this modeling ability, we resort to adding lightweight modules and reusing heavy dynamic operators in decoder layers.

Based on the above analysis, in this paper, we present a new paradigm, \ie, query-based object detector with cross-stage interaction, coined as StageInteractor.
This interaction of our method lies in two aspects: cross-stage label assignment and cross-stage dynamic filter reuse.
Specifically, during the label assignment, a \textit{cross-stage label assigner} is applied subsequent to the vanilla bipartite matching in each decoder layer. This assigner collects the results of bipartite matching across stages, and then reallocate proper training target labels for each object prediction according to the query index and a score. 
As for the forward pass, in each decoder layer, we \textit{reuse} the heavy dynamic operators in preceding stages and add lightweight modules to increase the modeling capacity at a relatively low cost.

Experiments show that our model with ResNet-50~\cite{resnet} as backbone can achieve 44.8~AP on MS COCO validation set under the simple setting of 100 queries, with 27.5~AP$_s$, 48.0~AP$_m$ and 61.3~AP$_l$ on small, medium and large object detection, respectively.
Equipped with $3\times$ training time, 300 queries and more data augmentation in line with other query-based detectors, our model can achieve 49.9 AP, 51.3 AP, and 52.7 AP with ResNet-101, ResNeXt-101-DCN~\cite{dcn1,deformableconv2}, and Swin-S~\cite{swin} as backbones, under the setting of single scale and single model testing.
Our model significantly outperforms the previous methods, as shown in~\cref{fig:convergence}, and it has become a new state-of-the-art query-based object detector.

\begin{figure*}
  \centering
  \includegraphics[width=0.99\linewidth]{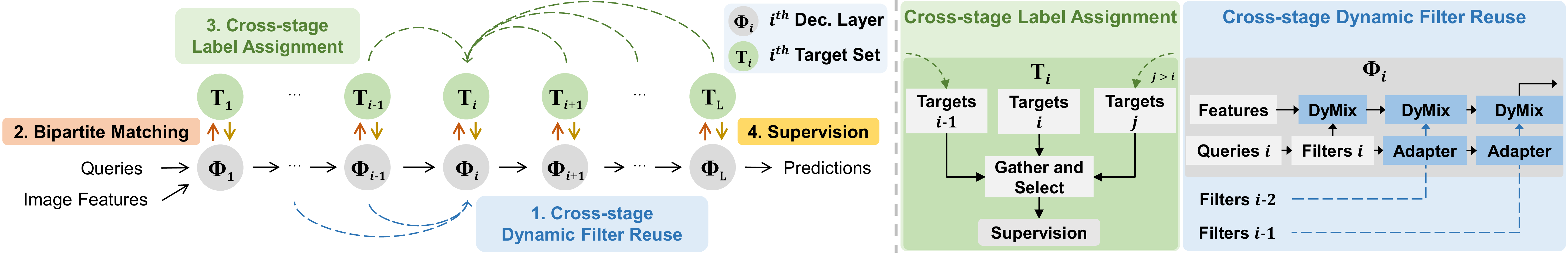}
   \caption{\textbf{Overview.} The cross-stage interaction incorporates two parts: cross-stage label assignment and cross-stage dynamic filter reuse. During the forward propagation, dynamic filters in each stage of decode layer are reused in the subsequent stages, \ie, we stack them with specific lightweight adapters to increase the depth of each decoder layer. As for the label assignment, our cross-stage label assigner gathers the results of bipartite matching across stages, and then selects proper target labels as supervision.}
   \label{fig:overview}
\end{figure*}

\section{Related Work}

DETR~\cite{detr} is an end-to-end query-based object detector without hand-crafted designs such as preset anchors and non-maximum suppression~(NMS), but it suffers from many problems, such as slow training convergence and unstable label assignment. To handle these challenges, a large quantity of works have been proposed. 
In this part, we divide these methods into two categories: modification of architectures and improvement of training procedures.

\textbf{Architecture.}
The vanilla DETR took a transformer encoder-decoder architecture~\cite{transformer} as the detection head, and used a set of object queries, content vectors, to encode the priors of object detection datasets.
In Deformable DETR~\cite{deformabledetr}, the attention operator~\cite{transformer} was replaced with multi-scale deformable attention module, and iterative bounding box refinement and a two-stage design which enables the detector to adaptively generate queries with a dense prior were also introduced.
In Sparse R-CNN~\cite{sparsercnn}, object query is decoupled into a content vector and an explicit bounding box, and image features are progressively aggregated into content vectors by ROIAlign~\cite{maskrcnn} with these boxes and dynamic convolution. Moreover, Sparse R-CNN is only a decoder architecture without any transformer encoder.
There were also many works like Conditional DETR~\cite{conditionaldetr}, DAB-DETR~\cite{dabdetr}, SMCA~\cite{smcadetr} and REGO~\cite{glimpsedetr} studying how to introduce spatial priors for proper features to accelerate convergence.
Adamixer~\cite{adamixer} improved Sparse R-CNN via deformable operators and spatial dynamic convolution to increase the adaptability.
There were also works like DDQ~\cite{DenseDistinctQuery} which combines the dense and sparse queries to guarantee high recall.
In this paper, we improve query-based object detector from a new perspective of the scalability~\cite{deepvit,xingbo_transformer,vit1000} of the decoder, \ie, we reuse dynamic operators among decoder layers to capture more diverse and complex representations.

\textbf{Training Procedure.}
In vanilla DETR, a set prediction loss~\cite{detr} is adopted for training.
Recently, many papers have analyzed how to accelerate the convergence of DETR via improved training procedures.
To verify whether the instability of Hungarian loss slow down the convergence, Sun~\etal~\cite{rethinkingdetr} utilized a matching distillation with a pre-trained DETR providing label assignment to train another model, and they found this instability only influenced the convergence in the early few epochs. 
DN-DETR~\cite{dndetr} presented a denoising training strategy where a set of additional noised ground-truth objects were passed through the decoder to reconstruct the corresponding raw objects.
DINO-DETR~\cite{dinodetr} improved DN-DETR via contrastive denoising training for hard negative samples.
Group DETR~\cite{groupdetr} introduced multiple groups of object queries for the global one-to-many label assignment during training, but maintained one-to-one label assignment in each group, and thus Group DETR could achieve the ability of duplicate removal with one group of queries.
Hybrid Matching~\cite{hybriddetr} was also proposed to combine one-to-one and one-to-many label assignment into one query-based object detector with a large number of queries, and it had three types of implementations: hybrid branch scheme (one-to-many matching for one group of queries, one-to-one matching for another), hybrid epoch scheme (one-to-many matching in early epochs, one-to-one matching in late epochs) and hybrid layer scheme (one-to-many matching in early layers, one-to-one in late layers).
Different from the previous methods, in this paper, we focus on the calibration of label assignment without adding additional queries. We collect training target labels across stages by a cross-stage label assigner, and then select proper targets to act as the supervision of each object prediction.

\section{Proposed Approach}
In this paper, we focus on the cross-stage interaction in query-based object detectors because it can well mitigate the misalignment between the decoders and supervisions in an object detector.
We first revisit the state-of-the-art query-based object detectors, especially AdaMixer~\cite{adamixer}, and then elaborate on our proposed cross-stage interaction.

\subsection{Preliminary on query-based object detectors}
\label{sec:revisit}

\begin{figure}
  \centering
  \includegraphics[width=1.\linewidth]{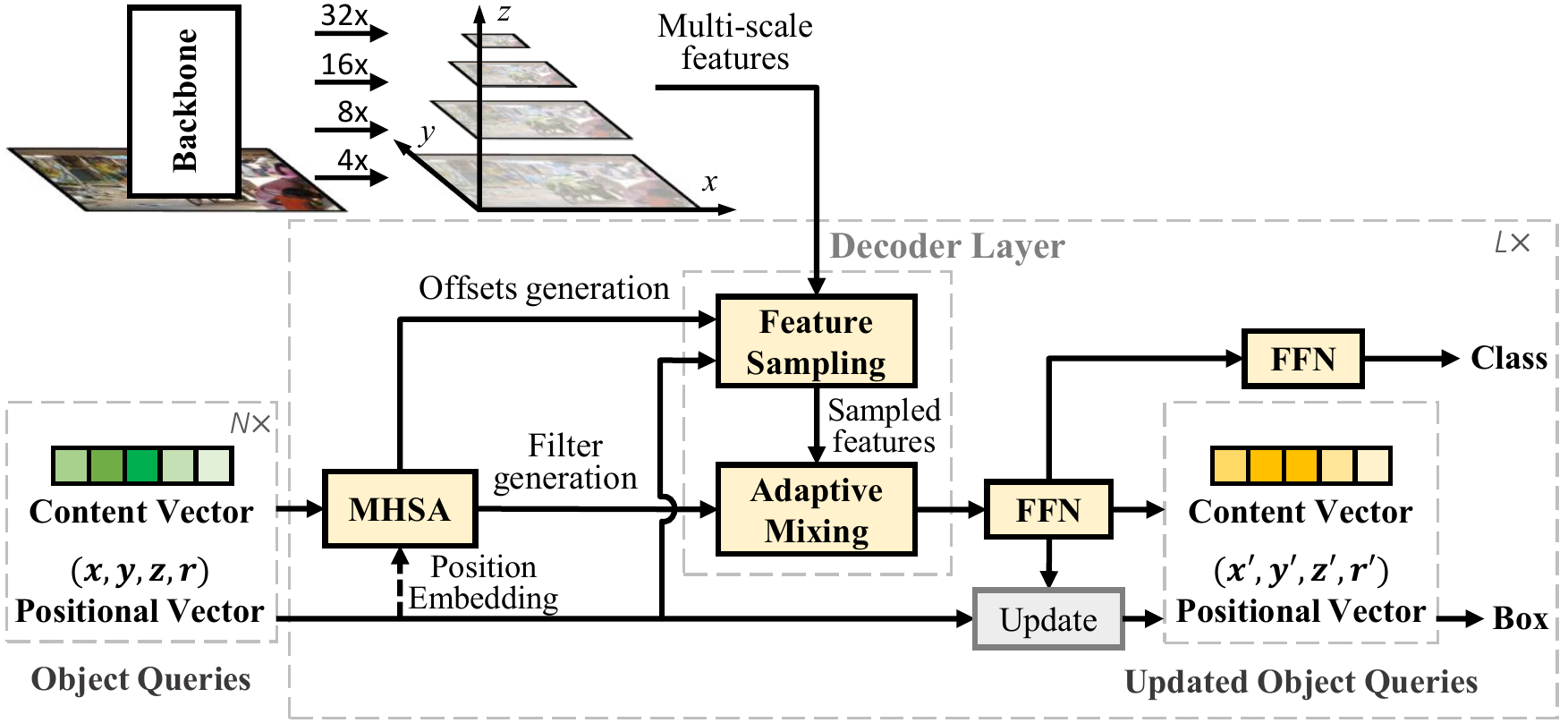}
   \caption{Overview of AdaMixer.}
  \label{fig:revisit_det}
\end{figure}

Generally, the architecture of query-based object detectors is composed by four parts: object queries, a backbone, a series of encoders and decoders.
Distinctively, Adamixer removes the encoders and still maintains the desired performance. As shown in~\cref{fig:revisit_det}, it consists of object queries, a backbone and a series of decoders.

\textbf{Object Query.}
The initial object queries is just a set of learnable embeddings. Recent object detectors~\cite{dabdetr,sparsercnn,adamixer} decompose them into content vectors and positional vectors.
The content vector is a vector $\mathbf{v} \in \mathbb{R}^{D}$.
The positional vector is presented in the format of box coordinates.
For example, in AdaMixer, it contains the information about the center point, scale and aspect ratio of an individual box.

\textbf{Decoder Layer.}
In a query-based detector, the decoder layers are stacked multiple times to form a cascade structure.
Each decoder layer is generally composed of three components: a multi-head self-attention~(MHSA), a dynamic interaction module, and feed-forward networks~(FFNs).
DETR-like models use the multi-head cross-attention for the dynamic interaction, while AdaMixer adopts a feature sampler and a dynamic mixing module, as shown in~\cref{fig:revisit_det}. 
The object queries are sequentially passed through these modules.
Specifically, the queries are \textit{first} processed by the multi-head self-attention module.
\textit{Then}, its outputs, the updated content vectors, together with the positional vectors are fed into the feature sampler.
In this sampler, each query is allocated with a unique group of regional multi-scale image features, \ie, \textit{\textbf{sampled features}}, by using the queries and bilinear interpolation.
\textit{Subsequently}, the adaptive mixing is performed on the sampled features with dynamic filters generated from the content vectors. Its outputs are aggregated into the vectors.
\textit{Last}, by means of FFNs, the queries updated by these modules can be decoded into object predictions, \ie, the relative scaling and offsets to the positional vectors~(bounding boxes), and the classification score vectors.
These updated queries also serve as inputs of the next stage.
Note that any query has and only has one corresponding prediction in every decoder layer.
Thus, we simply represent the predictions derived from the same initial query with one index, \ie, the \textit{\textbf{query index}}.

\textbf{Training and testing.}
In each decoder layer of a query-based detector, a bipartite matching is directly adopted on the ground-truth objects and the predictions. After some predictions match the ground-truth, these predictions are deemed as positive samples (assigned with foreground classification labels and supervision for localization) while others are the negative (assigned with the background label).
Focal loss~\cite{focalloss} serves as the classification loss and GIoU loss~\cite{giou} with $\ell_1 $ loss acts as the localization loss. 
During inference, only the outputs with high confidence from the last layer are used for evaluation.

\begin{figure}[t]
\centering
\begin{subfigure}{0.99\linewidth}
\centering
\includegraphics[width=0.99\linewidth]{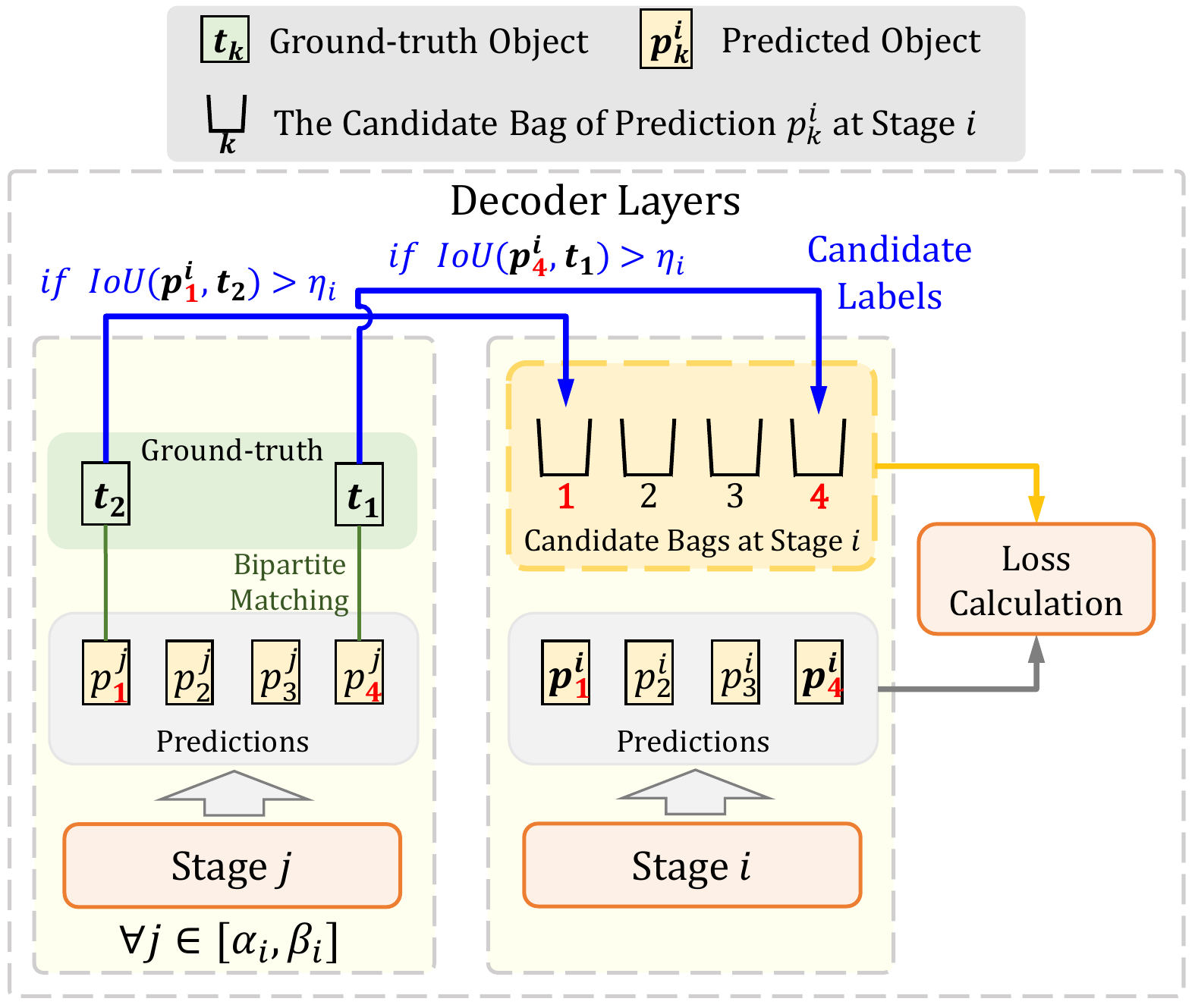}
\vspace{-.2em}
\caption{Given a stage $i$, we enumerate other decoder layers whose indexes are denoted as $j$, and select their assigned targets as the supervision for the stage $i$ according to~\cref{Equ:target_cond}.}
\vspace{1em}
\label{fig:csla}
\end{subfigure}

\begin{subfigure}{0.99\linewidth}
\centering
\includegraphics[width=0.5\linewidth]{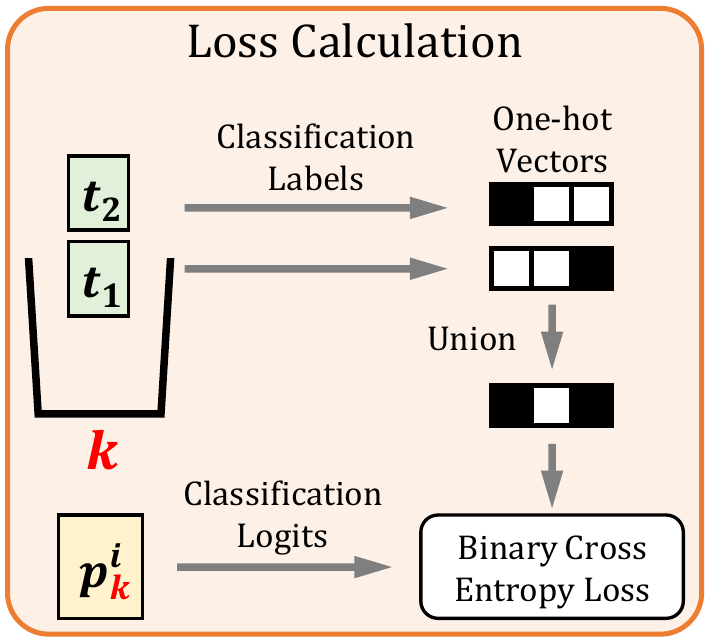}
\vspace{-.2em}
\caption{The elements in each candidate set are formed into the final supervision.}
\label{fig:csla_bce}
\end{subfigure} 
\vspace{-.5em}
\caption{The process of our cross-stage label assignment.}
\label{fig:csla_all}
\end{figure}

\subsection{Cross-stage Label Assignment}

To mitigate the training difficulty caused by bipartite matching in query-based object detectors, we propose a new \textit{cross-stage label assigner} to modify the results of the preceding bipartite matching. As depicted in~\cref{fig:overview}, our assigner first gathers these assignments across the \textit{stages} (\ie, decoder layers), and then selects appropriate training target class labels for each prediction. 

When gathering training targets, the cross-stage label assigner forces each prediction to only access the targets of predictions sharing the same \textit{query index} (defined in~\cref{sec:revisit}) with it.
The motivation behind this constraint is that the supervision of a single query may vary across stages, even though its predicted boxes across stages continuously cover the same ground-truth object, shown in~\cref{fig:la_example}\textcolor{red}{a}.
To alleviate this inconsistency for each query, we leverage its assigned targets provided by bipartite matching from multiple stages.
When introducing targets, we use a score to determine whether a match between a target and a prediction is suitable to be shared across stages.

\textbf{Algorithm.} 
Thanks to Focal loss~\cite{focalloss} with binary cross entropy loss in prevailing query-based object detectors~\cite{adamixer,sparsercnn,dndetr,dinodetr}, the multi-class classification task can be viewed as the binary classification on each category, and there is no need to consider a separate background category when training. More importantly, \textit{our cross-stage label assignment can be conducted on each category independently}. The specific algorithm is as follows:

Our cross-stage label assignment is performed between two stages with one type of condition.
\textit{First}, as shown in~\cref{fig:csla}, given a stage $i$ and its predictions $\{p^i_1, ..., p^i_N\}$, we traverse other stages with indexes in the range $[\alpha_i, \beta_i]$. Taking the stage $j$ as an example, we obtain its predictions $\{p^j_1, ..., p^j_N\}$.
\textit{Second}, we also create an empty candidate bag for each prediction of the stage $i$. We perform bipartite matching between the predictions of the stage $j$ and the ground-truth objects. Since this matching is an one-to-one label assignment, each ground-truth object corresponds to a prediction with a specific query index at the stage $j$. 
\textit{Third}, we align the predictions from these two stages according to the query indexes, and thus each ground-truth object also corresponds to a candidate bag.
We then select the training targets from the stage $j$ as the candidate supervisions according to the condition:
\begin{equation}
\begin{gathered}
\vartheta_i \big(q, t \big) \geq \eta_i ,
\end{gathered}
\label{Equ:target_cond}
\end{equation}
where $\vartheta_i \big(q, t \big)$ denotes a score between the query $q$ and the ground-truth object $t$ at the stage $i$. Here, the object $t$ needs to be assigned to query $q$ at stage $\Phi_j$ by the bipartite matching. $\eta_i$ denotes a threshold.
In practice, we follow the classical setting~\cite{fasterrcnn}, where the \textit{IoU} as the score and the threshold is set as 0.5.
Thus, if the \textit{IoU} is satisfied, the ground-truth objects are added to the corresponding candidate bags.
\textit{Finally}, after traversing all the stages in the range $[\alpha_i, \beta_i]$, we use the updated candidate bags to provide supervision for the stage $i$.
As shown in Figure~\ref{fig:csla_bce}, we convert the labels in each candidate bag into one-hot vectors, and then merge them into one vector. The resulting vector serves as classification supervision through the binary cross entropy loss (Focal loss~\cite{focalloss} in practice).

\begin{table}[t]
  \setlength{\tabcolsep}{5pt}
  \centering
  \begin{tabular}{l|ccc}
    \toprule
    Method & AP & AP$_{50}$ & AP$_{75}$ \\
    \midrule
    Baseline & 43.0 & 61.5 & 46.1 \\
    Hybrid Layer$^\dagger$ &  41.8 (\textcolor{blue}{-1.2}) & 60.5 (\textcolor{blue}{-1.0}) & 44.7 (\textcolor{blue}{-1.4}) \\ 
    Ours &  44.8 (\textcolor{blue}{+1.8})  &  63.0 (\textcolor{blue}{+1.5}) & 48.4 (\textcolor{blue}{+2.3})  \\
    \bottomrule
  \end{tabular}
  \vspace{-.5em}
  \caption{We reproduce ($\dagger$) the hybrid matching scheme~\cite{hybriddetr}, and compare it with our cross-stage label assigner with 100 queries.}
  \label{tab:hybridlayer}
\end{table}

\textbf{Discussion.}
Although our label assignment seems to have something in common with existing works like TSP-RCNN~\cite{rethinkingdetr} and Hybrid Layer Matching~\cite{hybriddetr}, the discrepancy between their design and ours cannot be ignored: 
(1)~in TSP-RCNN, first, the idea from Faster-R-CNN~\cite{fasterrcnn} is directly borrowed into the set prediction problem (\ie, a ground-truth object is only assigned to the proposal that shares an IoU score greater than 0.5 with it).
TSP-RCNN adopts such strategy for both the classification and localization. Differently, we can apply it only for the \textit{classification} task, shown as~\cref{tab:whynot_reg}.
Second, TSP-RCNN adopts such strategy with dense proposals for both the classification and localization. Differently, we apply it with \textit{sparse queries}.
(2)~for Hybrid Layer Matching, ground-truth objects are simply replicated in the first four stages, and then a bipartite matching is conducted. Differently, we do not modify the set of ground-truth objects. We gather results of the bipartite matching across stages, and then only select some targets as supervision.
Also, we observe that Hybrid Layer Matching is incompatible with the models given few queries.
In~\cref{tab:hybridlayer}, we implement it with the basic setting of our detector with 100 queries. The results show that it is totally infeasible under such a circumstance.
In a nutshell, our approach is greatly different from these methods.

\begin{figure} 
  \centering
  \includegraphics[width=0.99\linewidth]{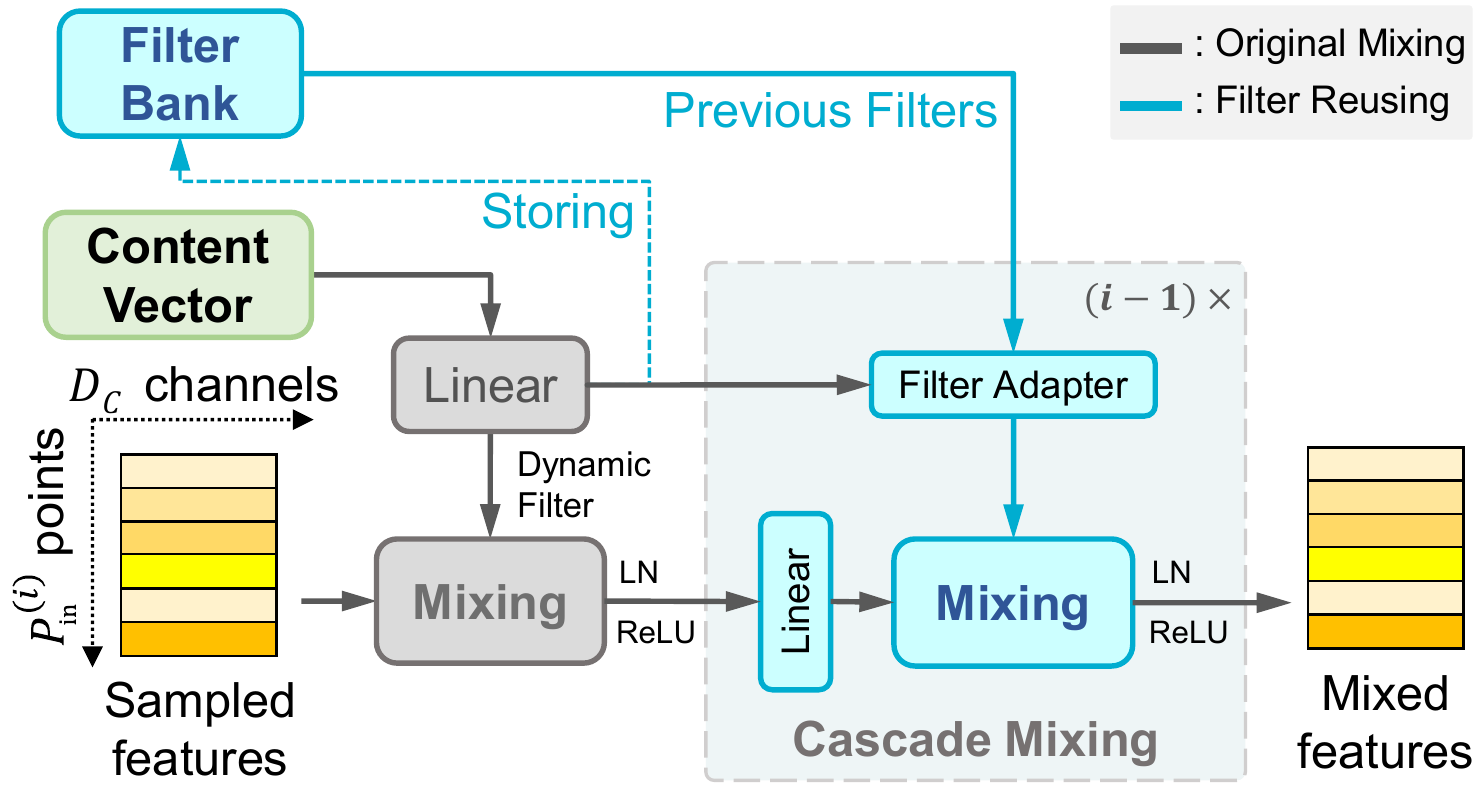}
   \caption{\textbf{The dynamic mixing with filter reusing on sampled features}. The content vectors dynamically generate filters through linear layers. These filters are used for mixing on sampled features, and then are stored into a dynamic filter bank for subsequent stages. Previous filters stored in the bank are also reused for mixing.}
  \label{fig:dynamic_overview}
\end{figure}

\subsection{Cross-stage Dynamic Filter Reuse}

The model capacity is essential for neural networks to capture complex representations~\cite{xingbo_transformer,deepvit}.
Since one of the major characteristics shared by query-based detectors is a series of decoder layers, we resort to modifying the decoder to improve this capacity.

A straightforward way of this modification is adding the attention-like operators.
The prevailing DETR-like detectors~\cite{dabdetr,dndetr,dinodetr} perform deformable cross-attention~\cite{deformabledetr} once along the spatial dimension in each stage, so directly adding this attention is feasible.
By contrast, other detectors~\cite{sparsercnn,adamixer} like AdaMixer perform more than one dynamic mixing at each stage, and they require a huge number of parameters~\cite{compare_detr_sparsercnn_data} to generate the corresponding dynamic filters.
Taking the channel mixing as an example, in each decoder layer, the specific process of generating a filter is as follows:
\begin{equation}
\begin{gathered}
\mathbf{M}_i = \mathbf{W}_0^{(i)} + \sum_{d=1}^{D} \mathbf{W}_d^{(i)} \mathbf{v}_{i,q,d} \in \mathbb{R}^{D_C \times D_C}, \\
\end{gathered}
\label{Equ:M}
\end{equation}
where $\mathbf{v}_{i,q,d}$ denotes the $d$-th element of the content vector of query $q$ at the stage $i$,
$\mathbf{W}_d^{(i)} \in \mathbb{R}^{D_C \times D_C}$ denotes a learnable weight matrix, and the number of such matrices in a stage is $D$,
$D_C$ denotes the channel dimension of the input features,
and $\mathbf{M}_i$ serves as the kernel for channel mixing~($1 \times 1$ convolution)~\cite{mlpmixer}.
Only the parameters used to generate a single filter have already been more than $D \times D_C^2$.
Thereby, it is impractical to directly stack more decoders given limited resources.
Fortunately, there are a large quantity of disposable dynamic filters in these cascade decoder layers, so these operators have the potential to be reused with \textit{lightweight modules} across stages.

As depicted in~\cref{fig:dynamic_overview}, we propose a cascade dynamic mixing module with a filter bank as an extension of the original channel mixing. The filters generated in each stage are stored in the filter bank for future use.
Given a stage, the \textit{filter adapters} update these stored filters with the ones from the current stage.
In each adapter, the process of updating filters is as follows:
\begin{equation}
\small
\begin{gathered}
\mathbf{w}_{j}^{(1)} = \sigma \big( \mathbf{W}_{j}^{(1)}\mathbf{v}_{i,q} \big) \in \mathbb{R}^{D_C} ,
\mathbf{w}_{j}^{(2)} = \sigma \big( \mathbf{W}_{j}^{(2)}\mathbf{v}_{i,q} \big) \in \mathbb{R}^{D_C},\\
\mathbf{M}_{j,i}' = \big( \mathbf{w}_{j}^{(1)} \cdot {\mathbf{w}_{j}^{(2)}}^\top \big) \odot \mathbf{M}_j + \big( \mathbf{1}-\mathbf{w}_{j}^{(1)} \cdot {\mathbf{w}_{j}^{(2)}}^\top \big) \odot \mathbf{M}_i, \\
\end{gathered}  
\label{Equ:adapters}
\end{equation}
where 
$\mathbf{1}$ denotes an all-ones matrix,
$\mathbf{W}_j^{(1)},\mathbf{W}_j^{(2)}$ denote learnable weights,
$\sigma(\cdot)$ denotes the sigmoid function, $\odot$ denotes the element-wise product, $\mathbf{M}_j$ denotes a previous filter, $j \in [\gamma_i, i)$, $\gamma_i$ is a lower bound, $\mathbf{M}_{j,i}'$ is used for subsequent modules, and the number of these updated filters is $\Delta \gamma_i = i - \gamma_i$.
We empirically find more filters can lead to more performance gain, so $\Delta \gamma_i = i-1$ is the best choice.
Note that this adapter only costs $2D_C^2$ parameters, which is more lightweight than the process of generating a filter from scratch as~\cref{Equ:M}.
Then, the updated filters are used in the cascade dynamic mixing. In this structure, since each dynamic mixing is followed by a layer normalization with an activation function, we insert a lightweight linear layer between every two dynamic mixing, consistent with~\cite{resnet}.

Moreover, we can modify the original spatial mixing with the dynamic filter reusing.
Unlike the channel mixing, the original dynamic spatial mixing costs a lot of computational resources on its ouput features due to its expansion~\cite{mobilenetv2} on spatial dimension.
To tackle this problem, we also adopt the bank to store the filters generated for spatial mixing.
Then, in each decoder layer, we update these filters with adapters and concatenate them with the current filter along the output dimension.
This new filter is used to perform spatial mixing only once, rather than iteratively.
This approach allows us to reduce the parameters of the filter generator of the spatial mixing, \ie, we employ the filters from preceding stages to replace a proportion of the filter from the current layer.

\begin{table}[t]
    \centering
    \renewcommand\arraystretch{1.0}
    \setlength{\tabcolsep}{2.8pt}
    \begin{tabular}{l|c|llllll}
        \toprule
        detector & AP  & AP$_{50}$ & AP$_{75}$ & AP$_{s}$ & AP$_m$ & AP$_l$ \\
        \midrule
         FCOS~\cite{fcos}  & 38.7 & 57.4 & 41.8 & 22.9 & 42.5 & 50.1 \\
         Cascade R-CNN~\cite{cascadercnn}  & 40.4 & 58.9 & 44.1 & 22.8 & 43.7 & 54.0 \\
         GFocalV2~\cite{gfl_v2}  & 41.1 & 58.8 & 44.9 & 23.5 & 44.9 & 53.3 \\
         BorderDet~\cite{borderdet} & 41.4 & 59.4 & 44.5 & 23.6 & 45.1 & 54.6 \\
         Dynamic Head~\cite{dynamichead_det}  & 42.6 & 60.1 & {46.4} & {26.1} & {46.8} & 56.0 \\
         DETR~\cite{detr}  & 20.0 & 36.2 & 19.3 & 6.0  & 20.5 & 32.2 \\
        Deform-DETR~\cite{deformabledetr}  & 35.1 & 53.6 & 37.7 & 18.2 & 38.5 & 48.7 \\
        Sparse R-CNN~\cite{sparsercnn}  & 37.9 & 56.0 & 40.5 & 20.7 & 40.0 & 53.5 \\
        AdaMixer~\cite{adamixer}   & {42.7} & {61.5} & {45.9} & {24.7} & {45.4} & {59.2}\\
        AdaMixer$^\dagger$~\cite{adamixer}  & {45.0} & {64.2} & {48.6} & {27.9} & {47.8} & {61.1} \\ 
        \midrule
        \textbf{StageInteractor}  & \textbf{44.8} & \textbf{63.0} & \textbf{{48.4}} & \textbf{{27.5}} & \textbf{{48.0}} & \textbf{61.3} \\ 
        \textbf{StageInteractor$^\dagger$}  & \textbf{46.9} & \textbf{65.2} & \textbf{{51.1}} & \textbf{{30.0}} & \textbf{{49.7}} & \textbf{62.3} \\ 
        \bottomrule
        \end{tabular}
    \vspace{-.5em}
    \caption{\textbf{$\mathbf{1}\times$ training scheme (12 epochs)} performance of various detectors on COCO \texttt{minival} set with ResNet-50. 100 object queries is the default setting in our method.
    $\dagger$ denotes 500 queries.
    }
    \vspace{-.5em}
    \label{table:onetimeschedule}
\end{table}

\section{Experiments}
\label{sec:experiments}

\subsection{Implementation Details}
The experiments are performed on the MS COCO~\cite{coco} object detection dataset, where the train2017 split and the val2017 split are for training and testing.
All of our experiments on AdaMixer are based on mmdetection codebase~\cite{mmdet}.
The experiments on DN-DETR and DINO are based on DETREX codebase~\cite{detrex}.
The convolutional neural networks~\cite{resnet,resnext} or vision trasnformers~\cite{van,swin,vit} can be taken as the backbone network.
8 RTX 2080ti with 11G GPU memory are enough to train our model with ResNet-50 and 100 queries. For larger backbones or more queries, we resort to 8 V100 with 32G.
The cross-stage label assignment is applied on each decoder layer.
The reuse of spatial dynamic filters do not perform on the first two decoder layers. 
The weighted sum of these losses are for training, and the loss weights are in line with those of AdaMixer in mmdetection codebase~\cite{mmdet} and those of DETRs in DETREX codebase~\cite{detrex}.
AdamW~\cite{adamw} is taken as the optimizer.

\begin{table*}
    \centering
    \renewcommand\arraystretch{1.0}
    \setlength{\tabcolsep}{4.8pt}
    \begin{tabular}{l|c|c|c|llllll}
        \toprule
        Detector & Backbone & {Encoder/FPN} & { Epochs}  & AP  & AP$_{50}$ & AP$_{75}$ & AP$_{s}$ & AP$_m$ & AP$_l$  \\
        \midrule
        DETR~\cite{detr} & ResNet-50-DC5 & TransformerEnc & 500  & 43.3 & 63.1 & 45.9 & 22.5 & 47.3 & 61.1  \\
        SMCA~\cite{smcadetr} & ResNet-50 & TransformerEnc & 50  & 43.7 & 63.6 & 47.2 & 24.2 & 47.0 & 60.4 \\
        Deformable DETR~\cite{deformabledetr} & ResNet-50 & DeformTransEnc & 50 & 43.8 & 62.6 & 47.7 & 26.4 & 47.1 & 58.0  \\
        Anchor DETR~\cite{anchordetr} & ResNet-50-DC5 & DecoupTransEnc & 50 & 44.2 & 64.7 & 47.5 & 24.7 & 48.2 & 60.6 \\ 
        Efficient DETR~\cite{efficientdetr} & ResNet-50 & DeformTransEnc & \textbf{36}  & 45.1 & 63.1 & 49.1 & 28.3 & 48.4 & 59.0 \\ 
        Conditional DETR~\cite{conditionaldetr} & ResNet-50-DC5 & TransformerEnc & 108  & 45.1 & 65.4 & 48.5 & 25.3 & 49.0 & {62.2}\\
        Sparse R-CNN~\cite{sparsercnn} & ResNet-50 & FPN & \textbf{36}  & 45.0 & 63.4 & 48.2 & 26.9 & 47.2 & 59.5  \\
        REGO~\cite{glimpsedetr} & ResNet-50 & DeformTransEnc & 50  & 47.6 & 66.8 & 51.6 & 29.6 & 50.6 & 62.3 \\
        DAB-D-DETR~\cite{dabdetr} & ResNet-50 & DeformTransEnc & 50  & 46.8 & 66.0 & 50.4 & 29.1 & 49.8 & 62.3 \\
        DN-DAB-D-DETR~\cite{dndetr} & ResNet-50 & DeformTransEnc & \textbf{12}  & 43.4 & 61.9 & 47.2 & 24.8 & 46.8 & 59.4 \\
        DN-DAB-D-DETR~\cite{dndetr} & ResNet-50 & DeformTransEnc & 50  & 48.6 & {67.4} & 52.7 & 31.0 & \textbf{52.0} & 63.7 \\
        AdaMixer~\cite{adamixer} & ResNet-50 & - & \textbf{12}   & 44.1 & 63.1 & 47.8 & 29.5 & 47.0 & 58.8  \\
        AdaMixer~\cite{adamixer} & ResNet-50 & - & \textbf{24}  & 46.7 & 65.9 & 50.5 & 29.7 & 49.7 & 61.5  \\ 
        AdaMixer~\cite{adamixer} & ResNet-50 & - & \textbf{36}  & {47.0} & {66.0} & {51.1} & {30.1} & {50.2} & {61.8}  \\
        StageInteractor & ResNet-50 & - & \textbf{12}  & {46.3}  & {64.3}  &  {50.6} & {29.8}  & {49.6}  &  {60.8}  \\
        StageInteractor & ResNet-50 & - & \textbf{24}   & {48.3}  & {66.6}  &  {52.9} & {31.7}  & {51.4}  &  {63.3}  \\
        StageInteractor & ResNet-50 & - & \textbf{36}   & \textbf{48.9} & \textbf{67.4} & \textbf{53.4 } & \textbf{31.7 } & {51.8} & \textbf{64.3}  \\
        StageInteractor$^*$ & ResNet-50 & - & \textbf{36}  & \underline{\textbf{50.8}} & \underline{\textbf{66.8}} & \underline{\textbf{55.9}} & \underline{\textbf{34.0}} & \underline{\textbf{54.6}} & \underline{\textbf{66.2}}  \\
        \midrule
        DETR~\cite{detr} & ResNet-101-DC5 & TransformerEnc & 500  & 44.9 & 64.7 & 47.7 & 23.7 & 49.5 & 62.3 \\
        SMCA~\cite{smcadetr} & ResNet-101 & TransformerEnc & 50  & 44.4 & 65.2 & 48.0 & 24.3 & 48.5 & 61.0 \\ 
        Efficient DETR~\cite{efficientdetr} & ResNet-101 & DeformTransEnc  & \textbf{36}  & 45.7 & 64.1 & 49.5 & 28.2 & 49.1 & 60.2 \\
        Conditional DETR~\cite{conditionaldetr} & ResNet-101-DC5 & TransformerEnc & 108   & 45.9 & 66.8 & 49.5 & 27.2 & 50.3 & 63.3 \\
        Sparse R-CNN~\cite{sparsercnn} & ResNet-101 & FPN & \textbf{36}  & 46.4 & 64.6 & 49.5 & 28.3 & 48.3 & 61.6 \\ 
        REGO~\cite{glimpsedetr} & ResNet-101 & DeformTransEnc & 50  & 48.5 & 67.0 & 52.4 & 29.5 & 52.0 & 64.4 \\ 
        AdaMixer~\cite{adamixer} & ResNet-101 & -  & \textbf{36}  & {48.0} & {67.0} & {52.4} & {30.0} & {51.2} & {63.7} \\ 
        StageInteractor & ResNet-101 & -  & \textbf{36}   & \textbf{49.9} & \textbf{68.6} & \textbf{54.6} & \textbf{33.0} & \textbf{53.6} & \textbf{65.4} \\ 
        \midrule
        REGO~\cite{glimpsedetr} & ResNeXt-101 & DeformTransEnc & 50  & 49.1 & 67.5 & 53.1 & 30.0 & 52.6 & 65.0  \\ 
        AdaMixer~\cite{adamixer} & ResNeXt-101-DCN & -  & \textbf{36}  & {49.5} & {68.9} & {53.9} & {31.3} & {52.3} & {66.3} \\ 
        StageInteractor & ResNeXt-101-DCN & -  & \textbf{36}  & \textbf{51.3} & \textbf{70.2} & \textbf{56.0} & \textbf{33.2} & \textbf{54.6} & \textbf{66.9} \\ 
        \midrule
        AdaMixer~\cite{adamixer} & Swin-S & -  & \textbf{36}  & {51.3} & {71.2} & {55.7} & {34.2} & {54.6} & {67.3} \\ 
        StageInteractor & Swin-S & -  & \textbf{36}   & \textbf{52.7} & \textbf{71.7} & \textbf{57.7} & \textbf{36.1} & \textbf{56.2} & \textbf{67.7} \\
        \bottomrule
        \end{tabular}
    \vspace{-.5em}
    \caption{The performance of various query-based detectors on MS COCO \texttt{minival} set with longer training scheme and single scale testing. The number of queries defaults to 300 in our method. $^*$ denotes 900 queries with more sampling points.
    }
    \label{tab:sota}
\end{table*}

\begin{figure}[t]
  \centering
  \includegraphics[width=0.99\linewidth]{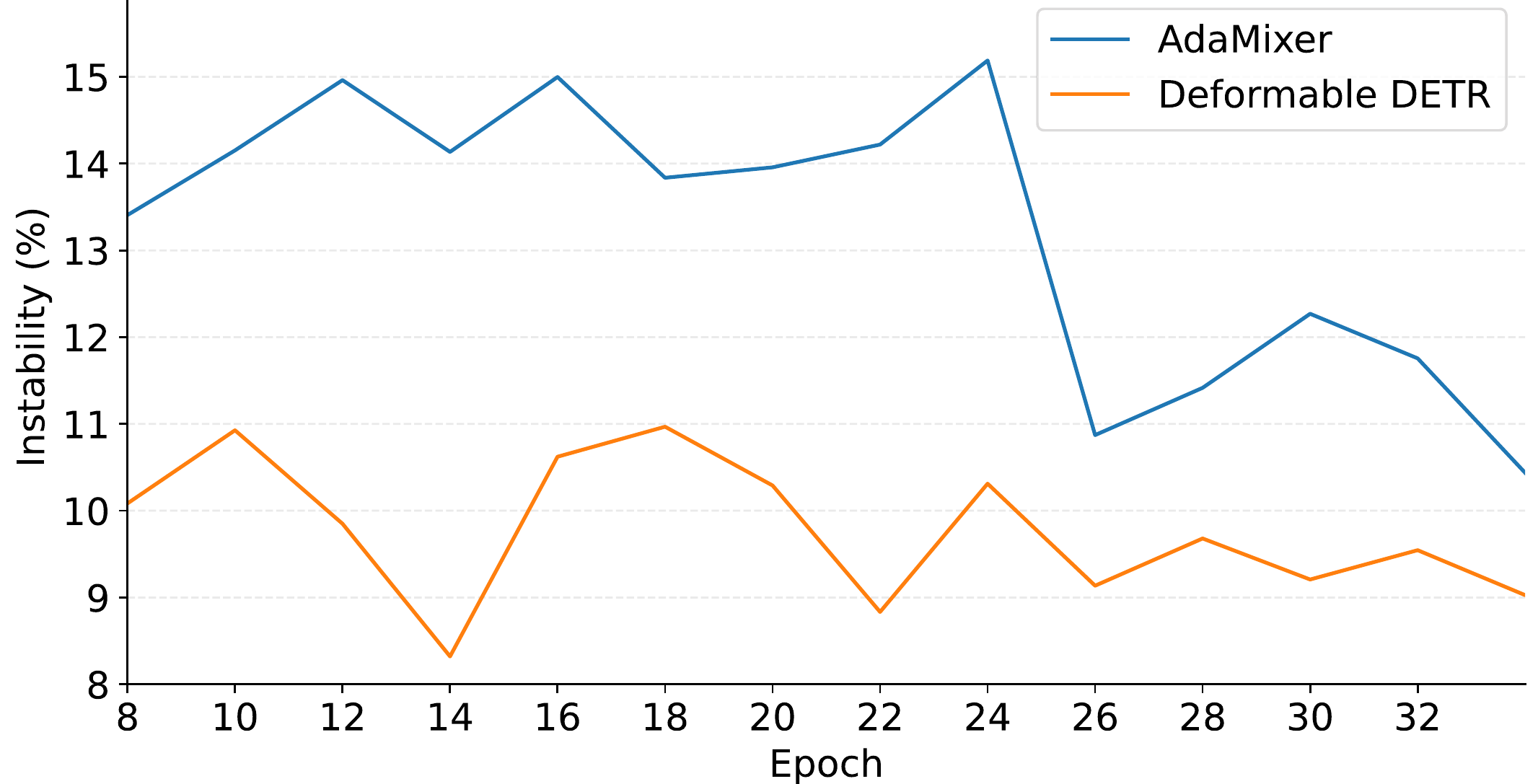}
  \vspace{-.5em}
   \caption{The instability of the assigned labels.}
  \vspace{-.5em}
   \label{fig:marginal_detr}
\end{figure}

\begin{table}[t]
\centering
\setlength{\tabcolsep}{5pt}
\begin{tabular}{l|c|c|ccc}
   \toprule
   Detector   & Epoch  & AP  & AP$_{50}$ & AP$_{75}$  \\
   \midrule
  DINO~\cite{dinodetr}   & 12 & 49.2 & 66.9 & 53.7 \\
  DINO + CSLA & 12 & {49.7}  & {67.0} & {54.1}  \\  
  DINO + CSLA + DCA & 12 & \textbf{50.0}  & \textbf{67.4} & \textbf{54.9}  \\  
  \midrule
  DINO~\cite{dinodetr}    & 24 & 50.6 & 68.5 & 55.5 \\
  DINO + CSLA & 24 & \textbf{51.0}  & 68.7 & 55.6\\ 
  DINO + CSLA + DCA & 24 & \textbf{51.3}  & \textbf{69.2} & \textbf{56.1}  \\ 
  \midrule
   Deform-DETR~\cite{deformabledetr} & 50  & 46.1 & 64.9 & 49.9  \\  
    Deform-DETR + CSLA & 50  & \textbf{46.8}  & \textbf{65.0} & \textbf{50.9} \\ 
  \midrule
   DN-DETR~\cite{dndetr}    & 50 & 44.7 & \textbf{65.3}  &  47.6 \\
   DN-DETR + CSLA & 50 & \textbf{45.5}   & \textbf{65.3} & \textbf{49.0}   \\ 
   \midrule
  H-DETR~\cite{hybriddetr}   & 12 & 48.6 & \textbf{66.3} & 53.2 \\  
  H-DETR + CSLA  & 12 & \textbf{49.2}  & \textbf{66.3} & \textbf{53.7} \\ 
  \bottomrule
\end{tabular}
\vspace{-.5em}
    \caption{The performance of DINO~\cite{dinodetr} on MS COCO \texttt{minival} set with ResNet-50. \texttt{CSLA}: cross-stage label assignment. \texttt{DCA}: dual cross-attention.
    }
    \label{tab:sota_dino}
\end{table}

\begin{table}[t]
\renewcommand\arraystretch{1.0}
\setlength{\tabcolsep}{5.0pt}
\centering
\begin{tabular}{l|c|ccc}
\toprule
Detector & Backbone   & AP & AP$_{50}$ & AP$_{75}$    \\
\midrule
DINO~\cite{dinodetr}  & Swin-B  & 55.8 & 74.4 & 60.7 \\
DINO~\cite{dinodetr} + CSLA & Swin-B  & \textbf{56.2} & 74.7 & 61.3 \\ 
\bottomrule
\end{tabular}
\vspace{-.5em}
\caption{Cross-stage label assignment on DINO~\cite{dinodetr} with Swin-B~\cite{swin} as backbone.}
\label{tab:swinb_dino}
\end{table}

\subsection{Comparison to State-of-the-Art Detectors}

Our model significantly outperforms the previous methods, shown in~\cref{fig:convergence}, and it has become a new state-of-the-art query-based object detector. 
As shown in~\cref{table:onetimeschedule}, with $\mathbf{1}\times$ training scheme and ResNet-50~\cite{resnet} as backbone, our detector outperforms various methods on COCO \texttt{minival} set~\cite{coco} with even with 100 queries. Our model with ResNet-50 as backbone can achieve 44.8~AP on MS COCO validation set under the basic setting of 100 object queries, with 27.5~AP$_s$, 48.0~AP$_m$ and 61.3~AP$_l$ on small, medium and large object detection, respectively.
When equppied with 500 queries, our detector performs better, and it can achieve 46.9 AP.
As depicted in~\cref{tab:sota}, equipped with $3\times$ training time, 300 queries and more data augmentation in line with other query-based object detectors, our model can achieve 48.9 AP, 49.9 AP, 51.3 AP, and 52.7 AP with ResNet-50, ResNet-101, ResNeXt-101-DCN~\cite{dcn1,deformableconv2}, and Swin-S~\cite{swin} as backbones, under the setting of single scale and single model testing.
Moreover, if we extend the quantity of queries into 900 and use tricks (\ie, adding more sampling points at each stage), our model can achieve 50.8~AP with ResNet-50. 

More importantly, our designs can be applied on DETR-like detectors such as DN-DETR~\cite{dndetr} and DINO~\cite{dinodetr}.
Unlike the cross-stage label assignment, our cross-stage filter reuse is tailored for AdaMixer~\cite{adamixer}.
To achieve larger model capacity,
we opt to simply double the cross-attention because this attention-like operation is more lightweight than the dynamic mixing.
In~\cref{tab:sota_dino}, our designs yield more than +0.5AP for all detectors.
To verify the effectiveness of our cross-stage label assignment on the larger backbones, we use DINO with Swin-B~\cite{swin} for experiments. As shown in~\cref{tab:swinb_dino}, our cross-stage label assignment can still yield +0.4AP on DINO with Swin-B as backbone with 12 epochs.

To find the reason for the lower performance gain on DETRs than that on AdaMixer, we present the instability of assigned labels (\ie, the probability of a ground-truth object transferring from one query to another across stages) as curves in~\cref{fig:marginal_detr}. This figure shows that Deformable-DETR is more stable, and we attribute this to its powerful transformer encoder which provides high-quality features for the decoder.

\subsection{Ablation Studies}
\label{sec:ablation}

Because the computational resource is limited, ResNet-50 is employed as our backbone network, the number of queries is set as 100, and $1\times$ training scheme is used for our ablation studies.
Due to the simplicity, we adopt AdaMixer as the baseline for ablation studies.

\textbf{The effectiveness of our proposed modules.} In~\cref{tab:add_baseline}, we conduct experiments to verify the effectiveness of our proposed modules. The results show that our modules lead to +2.2 AP gain, and each of them boosts the performance.

\textbf{The components of our label assigner.}
We verify how the components of our cross-stage label assigner influences the performance in~\cref{tab:abla_iou}.
In the first two lines, the results show that directly adding more supervisions based on the query index only brings marginal gains. According to the last two lines, we find that while using scores like IoU can boost the performance, using both IoU and the cross-stage labels leads to better results.

\textbf{The number of reused spatial filters.}
We explore the effectiveness of reusing spatial filters and the results are in~\cref{tab:Sstage_filter_offsets}.
We find a certain number of reused filters can bring a slight performance gain. This may result from the easier optimization brought by the fewer model parameters.

\textbf{Inference speed and training memory use.}
As depicted in~\cref{tab:speed}, with one TITAN XP and one batch size, the speed of our model is 11.6 img/s while that of the baseline is 13.5 img/s, \ie, only $\mathbf{1.16}\times$ slower.
When we set the batch size as 2 for training, the additional operation only costs about 0.3G GPU memory~(about \textbf{5}\%).

\textbf{The number of additional channel mixing.}
We conduct ablation studies on how many previous filters are required for the channel mixing, and report performance in~\cref{tab:Mstage_filter_offsets}.
We do experiments on the $\max \{ \Delta \gamma_i \}$. This means there are at most $\max \{ \Delta \gamma_i \}$ filters reused for the $i$-th stage. We find more filters can bring more performance gain. 
The results also show the scalability of the decoder layers in this framework.

\textbf{Selection of filters on mixing.}
We explore whether the previous filters with adapters are suitable for our new channel mixing.
As shown in~\cref{tab:pre_channel}, we find that using the adapters to fuse previous filters with the current ones is most suitable, which ensures both the adaptability of each stage and the diversity~\cite{deepvit} of filters.
More importantly, the existence of the additional dynamic mixing is necessary.

\textbf{Modifying the localization supervision.}
In our cross-stage label assigner, only classification labels are used for gathering, while the supervision for localization is unchanged.
Thus, we explore its influence by consistently updating the supervision of classification and localization, \ie selecting the ground-truth boxes that have the greatest IoU with predicted boxes across stages.
The results are shown in~\cref{tab:whynot_reg}, and we find that the performance under the two settings is very close, so we do not modify the localization part for simplicity.

\begin{table}[t]
\centering
\setlength{\tabcolsep}{4pt}
\begin{tabular}{cc|llllll}
   \toprule
   CSLA & Reuse  & AP  & AP$_{50}$ & AP$_{75}$ & AP$_{s}$ & AP$_m$ & AP$_l$ \\
    \midrule
    & & 42.6 & 61.4 & 45.7 & 24.4 & 45.7 & 58.2 \\
    & \checkmark & 43.0 & 61.5 & 46.1 & 25.0 & 45.7 & 59.1 \\
   \checkmark & & 44.1  & 62.3  & 47.6  &  25.5 & 47.5  & 60.3 \\ 
   \rowcolor{Gray}
   \checkmark & \checkmark  & \textbf{44.8} & \textbf{63.0} & \textbf{{48.4}} & \textbf{{27.5}} & \textbf{{48.0}} & \textbf{61.3} \\
     \bottomrule
\end{tabular}
  \vspace{-.5em}
  \caption{The effectiveness of our proposed modules. \texttt{CSLA}: cross-stage label assignment.}
  \label{tab:add_baseline}
\end{table}

\textbf{Application scope of the cross-stage label assigner.}
For the $i$-th stage, the application scope of cross-stage label assigner is $[\alpha_i, \beta_i]$.
In this part, we explore the appropriate values of $ \alpha_i $ and $ \beta_i $.
As shown in~\cref{tab:stage_la_offsets}, we find that the best setting is $[i-1, L] $.
Yet if too many previous ground-truth labels are included, like $\alpha_i = 1 $, this hinders the last decoder layer to learn how to remove duplicates.
Moreover, we conduct experiments to verify whether the cross-stage label assigner needs to be applied on all stages, because existing works~\cite{detr,hybriddetr} demonstrate that only early stages can be applied with one-to-many label assignment.
On the contrary, as shown in the last line of~\cref{tab:stage_la_offsets}, we find that our cross-stage label assigner can also be applied on the last two stages and this brings performance gain. We speculate that this module helps remove some inappropriate supervision caused by vanilla bipartite matching which has a shortage of constraining IoU, and this module provides some proper supervision from the previous stage.

\textbf{The threshold of IoU.} As shown in~\cref{tab:iou_thres}, we find that using a threshold of 0.5 is the best choice to select appropriate labels.

\begin{table}[t]
  \renewcommand\arraystretch{1.0}
  \setlength{\tabcolsep}{4pt}
  \centering
  \begin{tabular}{cc|llllll}
    \toprule
    Cross & Score  & AP  & AP$_{50}$ & AP$_{75}$ & AP$_{s}$ & AP$_m$ & AP$_l$ \\
    \midrule
     &  & 43.0 & 61.5 & 46.1 & 25.0 & 45.7 & 59.1 \\
    \checkmark  &   & 43.1 & 61.9 & 46.5 & 25.3 & 45.8 & 59.7 \\
    & \checkmark  & 44.0 & 62.1  & 47.6  & 26.0  & 47.6  & 59.3  \\
    \rowcolor{Gray}
    \checkmark & \checkmark  & \textbf{44.8} & \textbf{63.0} & \textbf{{48.4}} & \textbf{{27.5}} & \textbf{{48.0}} & \textbf{61.3} \\
    \bottomrule
  \end{tabular}
  \vspace{-.5em}
  \caption{The effectiveness of the components of the cross-stage label assigner. }
  \label{tab:abla_iou} 
\end{table}

\begin{table}[t]
  \renewcommand\arraystretch{1.0}
  \centering
  \begin{tabular}{c|llllll}
    \toprule
    $N_S$  & AP  & AP$_{50}$ & AP$_{75}$ & AP$_{s}$ & AP$_m$ & AP$_l$ \\
    \midrule
    0  & 44.5 & 62.5 & 48.1 & 27.4 & 48.0 & 60.6 \\  
    \rowcolor{Gray}
    1  & \textbf{44.8} & \textbf{63.0} & \textbf{{48.4}} & \textbf{{27.5}} & \textbf{{48.0}} & \textbf{61.3} \\
    3  &  44.4 & 62.4  & 48.0  &  26.2 &  47.4 & 60.6  \\  
    \bottomrule
  \end{tabular}
  \vspace{-.5em}
  \caption{The number of reused spatial filters.}
  \label{tab:Sstage_filter_offsets}
\end{table}

\begin{table}[t]
  \renewcommand\arraystretch{1.0}
  \centering
  \begin{tabular}{c|ccc}
    \toprule
     Method  & AP & Speed (img/s) & Memory (GB)\\ 
    \midrule
    Baseline  & 42.6 &  13.5 & 6.0 \\
    \rowcolor{Gray}
    Ours  & 44.8 & 11.6 & 6.3 \\
    \bottomrule
  \end{tabular}
  \vspace{-.5em}
  \caption{The inference speed and training GPU memory use.}
  \label{tab:speed}
\end{table}

\begin{table}[t]
  \renewcommand\arraystretch{1.0}
  \setlength{\tabcolsep}{4pt}
  \centering
  \begin{tabular}{c|llllll}
    \toprule
     $\max \{ \Delta \gamma_i \}$ & AP  & AP$_{50}$ & AP$_{75}$ & AP$_{s}$ & AP$_m$ & AP$_l$ \\
    \midrule
    0  & 44.1  & 62.3  & 47.6  &  25.5 & 47.5  & 60.3 \\
    1  & 44.2 & 62.3 & 47.6 & 25.8 & 47.5 & 60.6 \\
    2  & 44.2 & 62.3  &  47.9 & 26.1 &  48.0 & 60.4  \\
    3  & 44.6  & 62.8  & 48.3  & 26.7 &  47.5 &  60.7 \\ 
    \rowcolor{Gray}
    4   & \textbf{44.8} & \textbf{63.0} & \textbf{{48.4}} & \textbf{{27.5}} & \textbf{{48.0}} & \textbf{61.3} \\
    \bottomrule
  \end{tabular}
  \vspace{-.5em}
  \caption{The number of additional channel mixing.}
  \label{tab:Mstage_filter_offsets}
\end{table}

\begin{table}[t]
  \renewcommand\arraystretch{1.0}
  \centering
  \begin{tabular}{c|llllll}
    \toprule
    Filter   & AP  & AP$_{50}$ & AP$_{75}$ & AP$_{s}$ & AP$_m$ & AP$_l$ \\
    \midrule
     & 43.5 & 61.7 & 46.8 & 25.3 & 47.0 & 59.0 \\
    Pre & 44.2 & 61.9 & 47.6 & 26.1 & 47.6 & 60.0 \\ 
    Cur & 44.5 & 62.7 & 47.9 & 26.5 & 47.7 & 60.5 \\
    \rowcolor{Gray}
    Adp  & \textbf{44.8} & \textbf{63.0} & \textbf{{48.4}} & \textbf{{27.5}} & \textbf{{48.0}} & \textbf{61.3} \\
    \bottomrule
  \end{tabular}
  \vspace{-.5em}
  \caption{Usage of filters. \texttt{Pre}, \texttt{Cur}: directly using previous and current filters. \texttt{Adp}: using adapters.}
  \label{tab:pre_channel}
\end{table}

\begin{table}[t]
  \renewcommand\arraystretch{1.0}
  \centering
  \begin{tabular}{c|llllll}
    \toprule
    Loc.   & AP  & AP$_{50}$ & AP$_{75}$ & AP$_{s}$ & AP$_m$ & AP$_l$ \\
    \midrule
      & \textbf{44.8} & 62.8 & \textbf{48.4} & 26.7 & \textbf{48.2} & 60.7 \\
     \rowcolor{Gray}
     \checkmark  & \textbf{44.8} & \textbf{63.0} & \textbf{{48.4}} & \textbf{{27.5}} & {48.0} & \textbf{61.3} \\
    \bottomrule
  \end{tabular}
  \vspace{-.5em}
  \caption{Adding localization supervision in cross-stage label assigner.}
  \label{tab:whynot_reg}
\end{table}

\begin{table}[t]
  \renewcommand\arraystretch{1.0}
  \setlength{\tabcolsep}{2pt}
  \centering
  \begin{tabular}{ccc|llllll}
    \toprule
    Range of $i$ & $\alpha_i$ & $\beta_i$ & AP  & AP$_{50}$ & AP$_{75}$ & AP$_{s}$ & AP$_m$ & AP$_l$ \\
    \midrule
   ${[1, L]}$ & $ i$ & $ i$  & 44.0 & 62.1  & 47.6  & 26.0  & 47.6  & 59.3 \\
   ${[1, L]}$ & $ i-1$ & $ i$ & 44.1 & 62.2 & 47.7 & 25.9 & 47.5 & 59.8 \\
   ${[1, L]}$ &  $ i-1$ & $ i+1$  & 44.1 & 62.4  & 48.0  & 26.5 & 47.3  & 60.1 \\
   \rowcolor{Gray}
   ${[1, L]}$ &  $ i-1$ & $ L$   & \textbf{44.8} & \textbf{63.0} & \textbf{{48.4}} & \textbf{{27.5}} & \textbf{{48.0}} & \textbf{61.3} \\
   ${[1, L]}$ &  $ 1$ & $ L$  & 43.3 & 60.7  & 46.8  & 25.7 & 47.2  & 58.6 \\  
   ${[1, L-2]}$ &  $ i-1$ & $ L$ & 44.3 & 62.4 & 47.7 & 26.1 & 47.6 & 60.4 \\ 
    \bottomrule
  \end{tabular}
  \vspace{-.5em}
  \caption{The application scope of the cross-stage label assigner for each stage.}
  \label{tab:stage_la_offsets}
\end{table}

\begin{table}[t]
  \small
  \renewcommand\arraystretch{1.0}
  \setlength{\tabcolsep}{5pt}
  \centering
  \begin{tabular}{c|llllll}
  \toprule
   IoU thres   & AP  & AP$_{50}$ & AP$_{75}$ & AP$_{s}$ & AP$_m$ & AP$_l$ \\
  \midrule
    0.3 & 43.8 & 62.4 & 47.2 & 26.3 & 46.8 & 60.3 \\
    0.7 & 44.4 & 61.1 & 48.4 & 26.0 & 47.5 & 61.5 \\ 
    \rowcolor{Gray}
    0.5  & \textbf{44.8} & \textbf{63.0} & \textbf{{48.4}} & 27.5 & 48.0 & 61.3 \\
  \bottomrule
  \end{tabular}
  \vspace{-.5em}
  \caption{The threshold of IoU in cross-stage label assignment.}
  \label{tab:iou_thres}
\end{table}

\section{Conclusion}

In this paper, we have presented a new fast-converging query-based object detector with cross-stage interaction, termed as StageInteractor, to alleviate inconsistency between the one-to-one label assignment and improve the modeling capacity.
Our proposed cross-stage label assigner gathers assigned labels across stages, and then re-assigns proper labels to predictions in each stage.
We also accumulate dynamic filter across stages and reuse them with lightweight modules to increase the modeling capacity without introducing too many parameters. 
With these two unique designs, StageInteractor significantly outperforms the previous methods, and our proposed label assignment can boost the performance of various query-based object detectors.

\paragraph{\bf Acknowledgements.} {
This work is supported by National Key R$\&$D Program of China (No. 2022ZD0160900), National Natural Science Foundation of China (No. 62076119, No. 61921006), Fundamental Research Funds for the Central Universities (No. 020214380091, No. 020214380099), and Collaborative Innovation Center of Novel Software Technology and Industrialization.
The authors would like to thank Ziteng Gao for helpful discussions.
}

\appendix

\section{Spatial Mixing}

\begin{figure}[t]
  \centering
  \includegraphics[width=1.\linewidth]{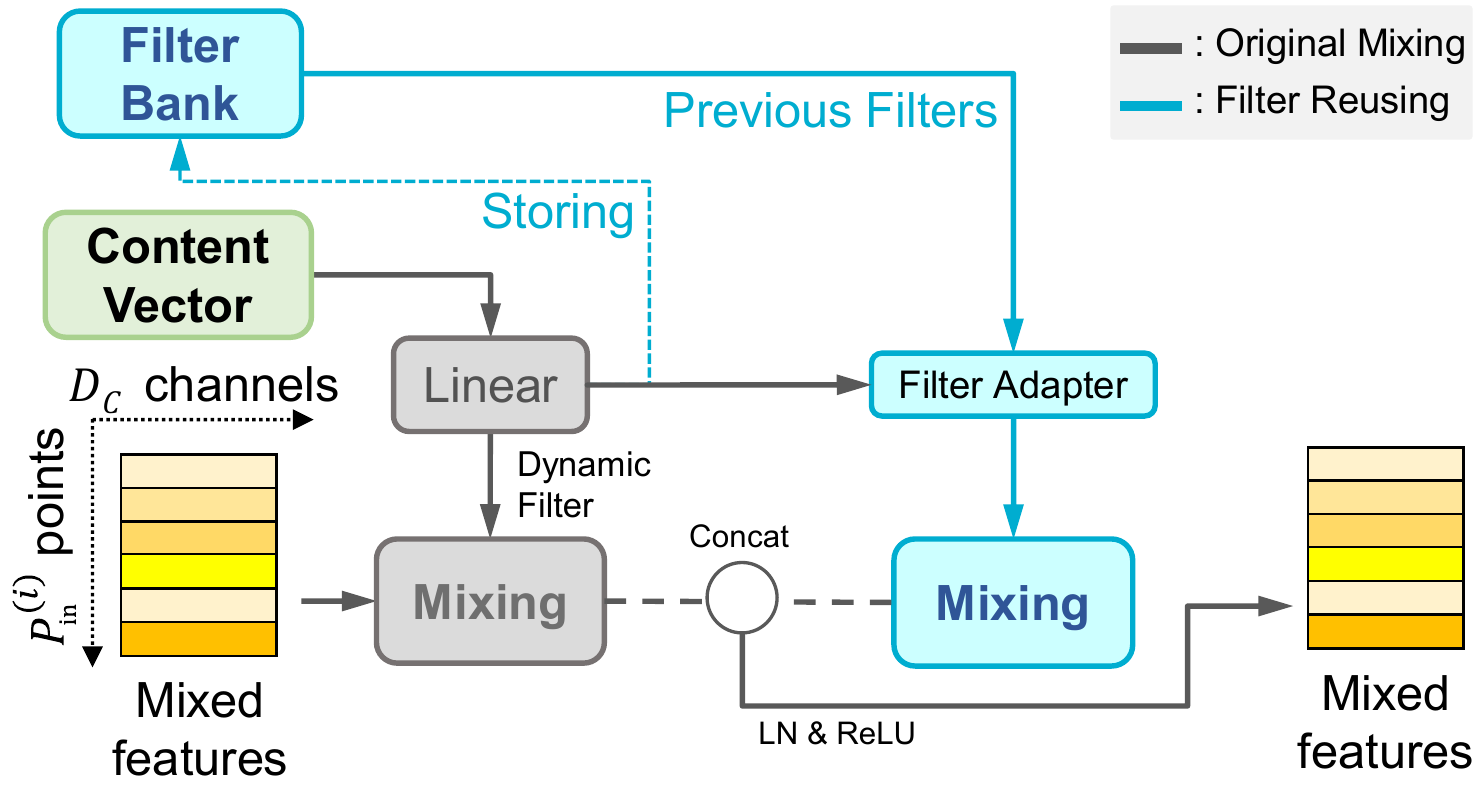}
  \vspace{-2.0em}
   \caption{The overview of our spatial dynamic mixing.}
   \label{fig:S}
\end{figure}

The process of applying filter reusing onto the spatial mixing is depicted in~\cref{fig:S}.
It is very similar to the filter reusing on channel mixing, but the reused filter is used in the combination with the generated filter rather than cascade mixing.
This combination is performed along the output dimension.

In the cascade mixing, the lightweight static linear layers are placed between the activation function and the dynamic mixing.
Apart from the static channel mixing, the linear layers can also achieve the \textit{efficient} spatial mixing, as shown in Code~\ref{code:dynamic_block}.
Specifically, we split the sampling points into $K$ groups, and perform affine transformation within and across groups, like~\cite{cyclemlp}. The parameters cost on this operation is $(K^2+(\frac{P^{(i)}_{\mathrm{in}}}{K})^2) \cdot D_C^2$.
Since the number of sampling points is set as the power of 2 and the number of spatial blocks $K$ is set close to the square root of the number of sampling points, we use the formula $K=2^{\lfloor \log_2 \sqrt{P^{(i)}_{\mathrm{in}}} \rfloor}$ for calculation. Thus, an upper-bound of the parameter cost is $O(3 P^{(i)}_{\mathrm{in}} D_C^2)$, and thus this module is still more lightweight than those related to dynamic filter generation.
\begin{Code} 
\centering
\begin{minipage}{1.0\linewidth}
\small
\begin{lstlisting}[language=python]
# K: spatial group size, P: the number of sampling points
# G: channel group size, Dc: channel dimension per group
# N: the number of queries
# I: the sampled feaures with shape (N*G, K, P//K, Dc)

I_1 = I.reshape(N*G*K, P//K*Dc)
I_2 = I.permute(0, 2, 1, 3).reshape(N*G*P//K, K*Dc)
I_1 = Linear(I_1).reshape(N*G, K, P//K, Dc)  
I_2 = Linear(I_2).reshape(N*G, P//K, K, Dc)
I = I + I_1 + I_2.permute(0, 2, 1, 3)
I = ChannelMixing(I)
\end{lstlisting}
\vspace{-1.5em}
\captionof{Code}{PyTorch-like Pseudocode for the static mixing.}
\label{code:dynamic_block}
\end{minipage}
\vspace{-1em}
\end{Code} 

\section{Feature Sampling}

For the feature sampling, according to~\cite{adamixer,deformabledetr,deformableconv1}, we first generate a set of sampling points via content vectors, and then use these points to capture the desired image features with bilinear interpolation.
Since the sampling points are organized into $K$ groups, the feature sampler is correspondingly designed to generate points in groups. The PyTorch-like Pseudo-code is illustrated in Code~\ref{code:sampler}.
Specifically, we first use the content vectors to generate two sets of offsets to the positional vectors by linear layers.
Then, the offsets is formed into the sampling points to extract features.

\begin{Code} 
\centering
\begin{minipage}{1.0\linewidth}
\small
\begin{lstlisting}[language=python]
# K: spatial group size, P: the number of sampling points
# G: channel group size, N: the number of queries
# v: the content vector, b: the positional vector
# F(): the bilinear interpolation on multi-scale image features
# Im: the multi-scale image feaures, I: the sampled feaures

xy = b[..., 0:2], z = b[..., 2:3], r = b[..., 3:4]

p_1 = Linear(v).reshape(N*G, K, 1, 3)  
p_2 = Linear(v).reshape(N*G, 1, P//K, 3)

dxy_1 = p_1[..., 0:2], dz_1 = p_1[..., 2:3]
dxy_2 = p_2[..., 0:2], dz_2 = p_2[..., 2:3]

p_xy = xy + 2**(z - 0.5*r) * (dxy_1 + 2**dz_1 * dxy_2)
p_z = z + dz_1 + dz_2
I = F(Im, p_xy, p_z)

\end{lstlisting}
\vspace{-1.5em}
\captionof{Code}{PyTorch-like Pseudocode of the sampler.}
\label{code:sampler}
\end{minipage}
\vspace{-1em}
\end{Code}

\section{Additional Ablation Studies}

\textbf{The modules in the cascade mixing.}
Both the reused heavy dynamic filters and the lightweight static linear layers are crucial to our method. As shown in~\cref{tab:cascade_mixing}, only when these two mixing approaches are combined can the large performance gain be achieved.
Moreover, as shown in~\cref{tab:ssm_shm}, we find that inserting static channel-spatial aggregation into the lightweight linear layers is more beneficial than solely performing channel or spatial mixing.

\begin{table}[t]
  \small
  \renewcommand\arraystretch{1.0}
  \setlength{\tabcolsep}{3pt}
  \centering
  \begin{tabular}{cc|llllll}
    \toprule
    Dynamic & Static  & AP  & AP$_{50}$ & AP$_{75}$ & AP$_{s}$ & AP$_m$ & AP$_l$ \\
    \midrule
     & & 44.1  & 62.3  & 47.6  &  25.5 & 47.5  & 60.3 \\ 
   & \checkmark & 43.5 & 61.7 & 46.8 & 25.3 & 47.0 & 59.0  \\
   \checkmark &    & 43.9 & 62.2 & 47.6 & 26.6 & 46.9 & 60.8  \\
  \checkmark &  \checkmark & \textbf{44.8} & \textbf{63.0} & \textbf{{48.4}} & \textbf{{27.5}} & \textbf{{48.0}} & \textbf{61.3} \\
    \bottomrule
  \end{tabular}
  \vspace{-.5em}
  \caption{The modules in the cascade mixing.}
  \label{tab:cascade_mixing}
\end{table}

\begin{table}[t]
  \small
  \renewcommand\arraystretch{1.0}
  \setlength{\tabcolsep}{3pt}
  \centering
  \begin{tabular}{c|llllll}
    \toprule
    Static Mixing  & AP  & AP$_{50}$ & AP$_{75}$ & AP$_{s}$ & AP$_m$ & AP$_l$ \\
    \midrule
      & 43.9 & 62.2 & 47.6 & 26.6 & 46.9 & 60.8 \\ 
    Channel  & 44.0 & 62.3 & 47.8 & 26.2 & 47.0 & 61.1 \\
    Spatial  & 43.7 & 61.8 & 47.2 & 25.7 & 47.0 & 59.9 \\
    Channel-spatial  & \textbf{44.8} & \textbf{63.0} & \textbf{{48.4}} & \textbf{{27.5}} & \textbf{{48.0}} & \textbf{61.3} \\
    \bottomrule
  \end{tabular}
  \vspace{-.5em}
  \caption{The type of static mixing in our detector.}
  \label{tab:ssm_shm}
\end{table}

\textbf{Feature sampling.}
Different from the feature sampling in~\cite{adamixer}, the sampler in our detector is required to generate points in groups. Therefore, in this part, we explore whether the original feature sampling method~(\ie directly generating all sampling points) is feasible. 
As shown in~\cref{tab:sampling}, we report the results of our detector with vanilla feature sampling in the first line.
Compared to the first line, the results in the last line show that our sampling is more compatible with our detector than 3D feature sampling.
To find whether the weight initialization of 3D feature sampler causes this phenomenon, we modify the initialization of sampler so that its outputs at the first iteration are identical with the our sampler, and report the corresponding performance in the second line.
The results are still worse than our sampling. Therefore, we speculate that our two-stage sampler is consistent with our dynamic mixing, thereby boosting the performance.

\begin{table}[t]
  \small
  \renewcommand\arraystretch{1.0}
  \setlength{\tabcolsep}{5.0pt}
  \centering
  \begin{tabular}{c|llllll}
    \toprule
    Sampling  & AP  & AP$_{50}$ & AP$_{75}$ & AP$_{s}$ & AP$_m$ & AP$_l$ \\
    \midrule
     Vanilla  & 43.7 & 62.1 & 47.2 & 25.6 & 47.4 & 59.7 \\  
    Group init.  & 44.1 & 62.3 & 47.9 & 26.2 & 47.0 & 60.8 \\  
    Ours  & \textbf{44.8} & \textbf{63.0} & \textbf{{48.4}} & \textbf{{27.5}} & \textbf{{48.0}} & \textbf{61.3} \\
    \bottomrule
  \end{tabular}
  \vspace{-.5em}
  \caption{The type of feature sampling. \texttt{Vanilla} denotes the original feature sampling~\cite{adamixer}. \texttt{Group-init.} means the group-wise initialization on the original sampler.}
  \label{tab:sampling}
\end{table}

\textbf{More sampling points in the first stage.}
As shown in~\cref{tab:more_points}, we conduct ablation studies on adding more sampling points of the first stage. The motivation is to use more points to cover the whole image as much as possible, enlarging the receptive field. Both of our baseline and our method can get slight benefits from more sampling points.

\begin{table}[t]
  \small
  \renewcommand\arraystretch{1.0}
  \setlength{\tabcolsep}{4pt}
  \centering
  \begin{tabular}{c|c|llllll}
  \toprule
  StageInter & $P^{(1)}_{\mathrm{in}}$ & AP  & AP$_{50}$ & AP$_{75}$ & AP$_{s}$ & AP$_m$ & AP$_l$ \\
  \midrule
   & 32  & 42.5 & 61.4 & 45.7 & 25.0 & 45.1 & 58.2 \\ 
   & 64  & 42.6 & 61.4 & 45.7 & 24.4 & 45.7 & 58.2 \\
   \midrule
  \checkmark & 32  & 44.6 & 62.6 & 48.3 & 26.2 & \textbf{48.1} & 61.1 \\ 
  \checkmark & 64   & \textbf{44.8} & \textbf{63.0} & \textbf{{48.4}} & \textbf{{27.5}} & 48.0 & \textbf{61.3} \\ 
  \bottomrule
  \end{tabular}
  \vspace{-.5em}
  \caption{The number of sampling points at the first stage.}
  \label{tab:more_points}
\end{table}

\section{Duplicate Removal}

According to~\cite{what_spz}, the strict one-to-one label assignment can ensure the object detector to have the ability to remove duplicate predictions.
However, our cross-stage label assigner actually does not strictly follow one-to-one matching even in the last few stages, \ie, it has the potential to assign one ground-truth object to multiple predicted boxes on the classification task.
Therefore, we explore whether our label assigner influences the performance of duplicate removal in query-based object detectors. 
As shown in~\cref{tab:nms}, the results show that the performance our detector is relatively stable on AP with or without NMS.
We consider this is because the coordinates of the most predicted boxes in the last few stages change little, and the operation of \textit{gathering-and-selecting} labels in our assigner is performed adaptively.

\begin{table}[t]
  \small 
  \centering
  \begin{tabular}{cc|cc}
    \toprule
    NMS &  Threshold  & AP  & AP$_{75}$  \\
    \midrule
    \checkmark & 0.5  & 44.1 (\textcolor{blue}{-0.7})  & 47.1 \\  
    \checkmark & 0.75  & 44.8 (\textcolor{blue}{+0.0})  & 48.4 \\
    \checkmark & 0.9  & 44.8 (\textcolor{blue}{+0.0})  & 48.5  \\
     - & - & {44.8}  & {48.4} \\ 
    \bottomrule
  \end{tabular}
  \vspace{-.5em}
  \caption{The performance of our StageInteractor with or without NMS.}
  \label{tab:nms}
\end{table}

\section{MS COCO Test}

As shown in~\cref{tab:coco_test}, we report the performance of StageInteractor on COCO \texttt{test-dev} set. Here, the performance is evaluated with the same models that are used for the comparison with other state-of-the-art query-based detectors.
Because the labels of COCO \texttt{test-dev} set are not publicly available, so the evaluation is performed on the online server.

\begin{table}[t]
  \small
  \renewcommand\arraystretch{1.0}
  \setlength{\tabcolsep}{4pt}
  \centering
  \begin{tabular}{c|llllll}
    \toprule
    Backbone  & AP  & AP$_{50}$ & AP$_{75}$ & AP$_{s}$ & AP$_m$ & AP$_l$ \\
    \midrule
    ResNet-50  & 49.0 & 67.4 & 53.7 & 30.2 & 51.7 & 62.3 \\  
    ResNet-101  & 50.4 & 68.8 & 55.2 & 31.0 & 53.1 & 64.1 \\  
    ResNeXt-101-DCN & 51.3 & 70.1 & 56.0 & 32.1 & 53.9 & 65.2 \\
    Swin-S  & 52.7 & 71.8 & 57.7 & 33.3 & 55.1 & 67.1 \\ 
    \bottomrule
  \end{tabular}
  \vspace{-.5em}
  \caption{The performance of StageInteractor on COCO \texttt{test-dev} set with 300 queries, 36 training epochs and single model single scale testing.}
  \label{tab:coco_test}
\end{table}

\begin{figure*}
  \centering
  \begin{subfigure}{0.33\linewidth}
    \includegraphics[width=1.\linewidth]{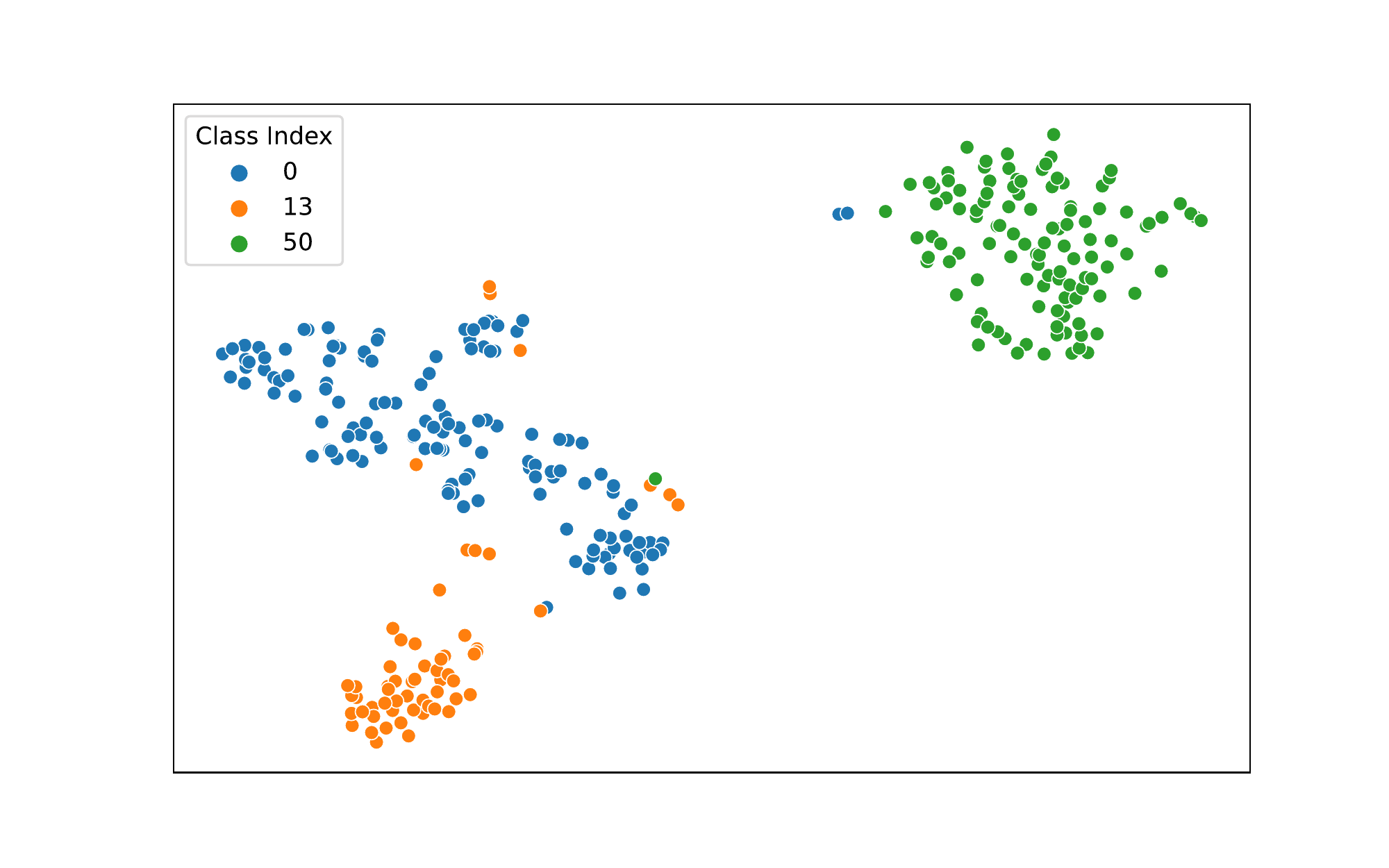}
    \caption{Baseline + CSLA.}
    \label{fig:tsne_noprefilter}
  \end{subfigure}
  \hfill
  \begin{subfigure}{0.33\linewidth}
    \includegraphics[width=1.\linewidth]{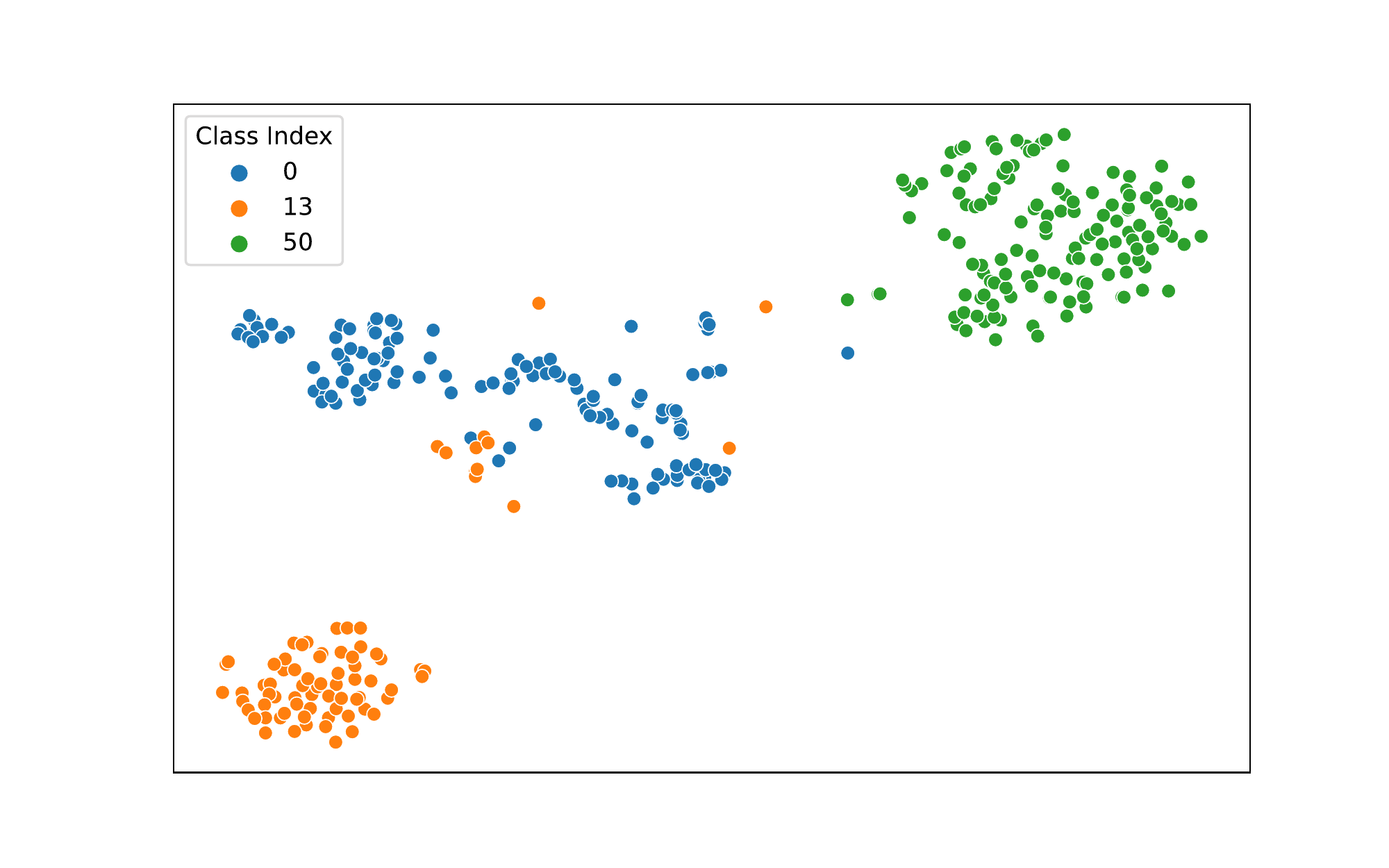}
    \caption{Baseline + Reusing.}
    \label{fig:tsne_nola}
  \end{subfigure}
  \hfill
  \begin{subfigure}{0.33\linewidth}
    \includegraphics[width=1.\linewidth]{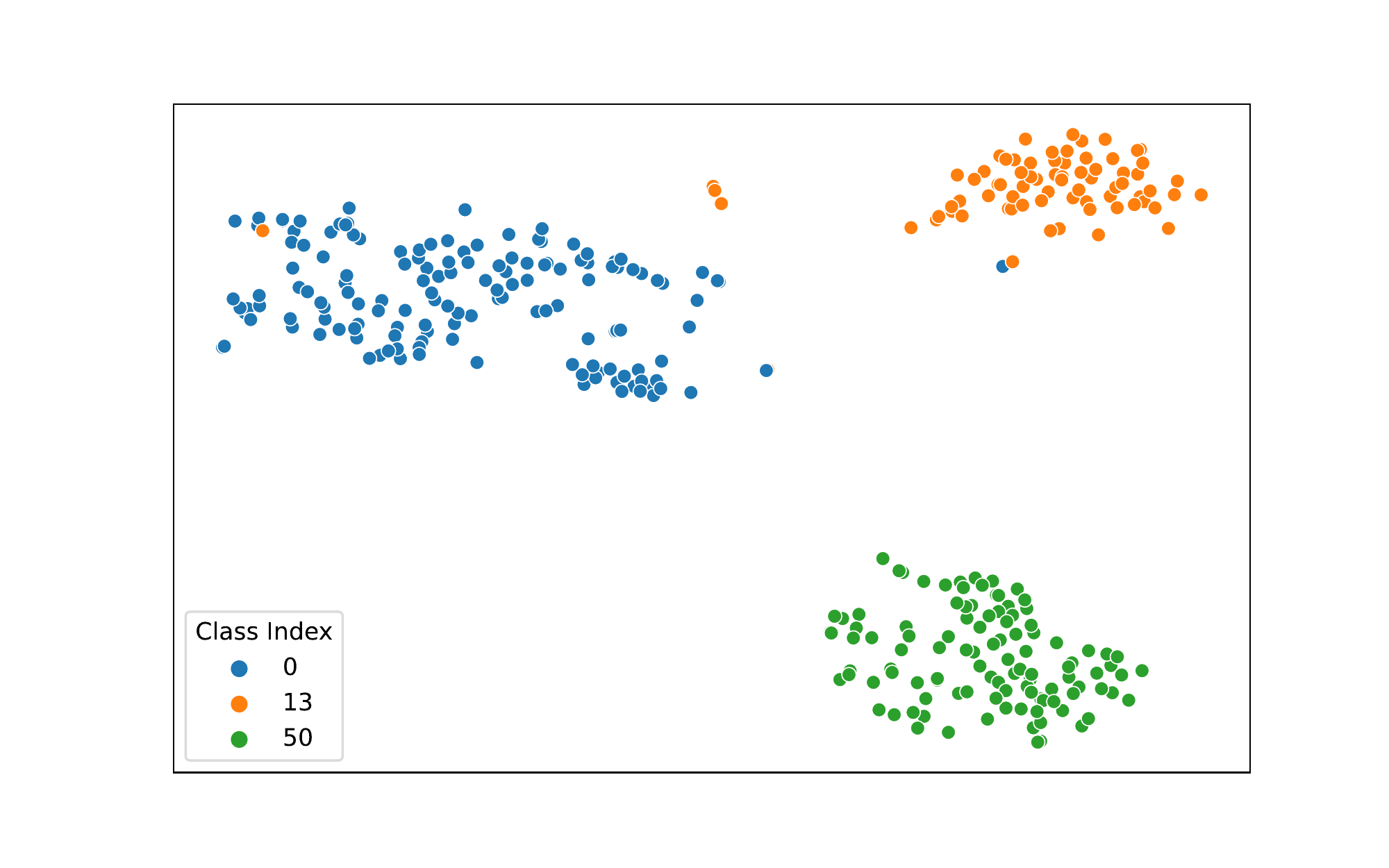}
    \caption{StageInteractor.}
    \label{fig:tsne_sota}
  \end{subfigure}
  \caption{t-SNE~\cite{tsne} visualization of features of each query on MS~COCO dataset~\cite{coco} learned by various model variants. Each point denotes a feature vector, and colors denote different categories. \texttt{CSLA}: cross-stage label assignment.}
  \label{fig:tsne}
\end{figure*}

\section{Analysis about dynamic channel mixing}

In vanilla AdaMixer~\cite{adamixer}, the FLOPs for the channel mixing is $B \times N \times G \times P^{(i)}_{\mathrm{in}} \times (2 D_C - 1) \times D_C $, whereas the FLOPs for generating a dynamic channel filter is $B \times N \times G \times (2D-1) \times D_C \times D_C $.
Therefore, the ratio between these two FLOPs is:
\begin{equation}
\frac{B \times N \times G \times (2D-1) \times D_C \times D_C }{B \times N \times G \times P^{(i)}_{\mathrm{in}} \times (2 D_C - 1) \times D_C } \approx  8
\end{equation}
Therefore, generating channel filters consumes more computational costs than performing channel mixing.

\section{Qualitative Analysis}

To verify the discriminability of our detector, we use t-SNE~\cite{tsne}  visualization for the query features in various models.
As depicted in~\cref{fig:tsne}, we select some representative categories with corresponding features to show the effectiveness of our structures.
Compared with~\cref{fig:tsne_noprefilter} and~\cref{fig:tsne_nola}, the distance between each group of categories is wider in~\cref{fig:tsne_sota}, and the points are more separate.

\section{Limitation}
Although our two cross-stage structures are effective, their designs are simple. In the future, we hope the following topics could be explored in the future:
(1)~the optimal designs of each decoder layer in a query-based detector;
(2)~more elaborate and effective cross-stage interactions;
(3)~the theoretical properties and the essence of the cascade structures.

\section{Societal Impact}
Object detection is a classical vision task and we adopt the open dataset: MS~COCO~\cite{coco}, so there is no negative social impact if the method is used properly.

\section{Model Implementation Details}

\textbf{Hyper-parameters.}
The cross-stage label assignment is performed on each stage, and its application scope is $[i-1, L]$.
The threshold for selecting labels is set to 0.5. 
The reuse of dynamic filters do not perform on the first two decoder layers, and in other stages, all the generated filters for channel mixing are reused.
The number of spatial blocks $K$ is set close to the square root of the number of sampling points, \ie, we use the formula $K=2^{\lfloor \log_2 \sqrt{P^{(i)}_{\mathrm{in}}} \rfloor}$ for calculation.
Other parameters of our model are in line with~\cite{adamixer}.
Other parameters of our model are in line with the vanilla AdaMixer~\cite{adamixer} and DETRs~\cite{detrex}.
 
\textbf{Initialization.}
Following~\cite{adamixer},
the initial weights of linear layers generating dynamic filters are set to zero, and the biases of these linear layers are initialized as expected.
The initial weights of linear layers in the feature sampler are also set to zero, and the biases of these linear layers are initialized as follows:~(1)~the bias corresponding to $\mathrm{dxy\_1}$ in Code~\ref{code:sampler} is uniformly initialized within {[-0.5, 0.5]}. (2)~the one corresponding to $\mathrm{dxy\_2}$ is uniformly initialized within {[-$\frac{0.5}{\sqrt{2}}$ , $\frac{0.5}{\sqrt{2}}$]}. (3)~The parts corresponding to the $\mathrm{dz\_1}$ and $\mathrm{dz\_2}$ are initialized as zeros.
The initialization of other modules are set following~\cite{adamixer,detrex}.

{\small
\bibliographystyle{ieee_fullname}
\bibliography{egbib}
}

\end{document}